\def\year{2021}\relax
\documentclass[letterpaper]{article} 
\usepackage{aaai21}  
\usepackage{times}  
\usepackage{helvet} 
\usepackage{courier}  
\usepackage[hyphens]{url}  
\usepackage{graphicx} 
\usepackage{natbib}  
\usepackage{caption} 
\usepackage{placeins}
\usepackage{stfloats}

\usepackage{graphicx}
\usepackage{booktabs,multirow} 
\usepackage{siunitx}
\usepackage{amsmath,amssymb}
\usepackage{cases}
\usepackage[dvipsnames]{xcolor}
\usepackage{graphicx}
\usepackage{bm}
\usepackage{subcaption}
\usepackage[export]{adjustbox}
\usepackage{makecell}
\usepackage{tikz}
\usetikzlibrary{arrows, arrows.meta, shapes, calc, intersections, positioning, decorations.pathreplacing, fit, backgrounds, patterns}

\usepackage[hyphens]{url}

\usepackage[noabbrev,nameinlink]{cleveref}

\newcommand{\rebuttal}[1]{#1}

\usepackage{etoolbox}           
\renewcommand{\bfseries}{\fontseries{b}\selectfont} 
\robustify\bfseries             
\newrobustcmd{\B}{\bfseries}

\newcommand{\x}{\mathbf{x}}
\newcommand{\y}{\mathbf{y}}
\newcommand{\z}{\mathbf{z}}

\DeclareMathSymbol{\sminus}{\mathbin}{AMSa}{"39}

\definecolor{vll-orange}{HTML}{E37238}
\definecolor{vll-green}{HTML}{96BF0D}
\definecolor{vll-dark}{HTML}{464646}
\definecolor{vll-light}{HTML}{757575}
\definecolor{mpl-blue}{rgb}{0.21568627450980393, 0.49411764705882355, 0.7215686274509804}

\pgfdeclarelayer{back}
\pgfdeclarelayer{background}
\pgfsetlayers{back,background,main}

\def\drawfill#1;{
  \fill[white] #1;
\begin{pgfonlayer}{back}
  \draw[line width=2pt] #1;
\end{pgfonlayer}}

\usepackage{listings}
\definecolor{codegray}{rgb}{0.4,0.4,0.4}
\definecolor{backcolour}{rgb}{0.97,0.97,0.97}
\lstdefinestyle{mystyle}{
    backgroundcolor=\color{backcolour},
    keywordstyle=\bf\color{black!75},
    numberstyle=\tiny\color{codegray},
    basicstyle=\ttfamily\scriptsize,
    breakatwhitespace=false,
    breaklines=true,
    captionpos=b,
    keepspaces=true,
    numbers=left,
    numbersep=3pt,
    showspaces=false,
    showstringspaces=false,
    showtabs=false,
    tabsize=2
}
\usepackage[switch]{lineno}
\let\linenumbers\nolinenumbers\nolinenumbers

\title{HINT: Hierarchical Invertible Neural Transport \\ for Density Estimation and Bayesian Inference}
\author {
    Jakob Kruse\thanks{equal contribution.}\textsuperscript{\rm 1}, \;
    Gianluca Detommaso$^\ast$\textsuperscript{\rm 2}, \;
    Ullrich K\"othe \textsuperscript{\rm 1}, \;
    Robert Scheichl \textsuperscript{\rm 1} \\
}
\affiliations {
    \textsuperscript{\rm 1} Heidelberg University, \;
    \textsuperscript{\rm 2} Amazon.com \\
    jakob.kruse@iwr.uni-heidelberg.de
}

\begin{document}
\linenumbers

\maketitle

\begin{abstract}

Many recent invertible neural architectures are based on coupling block designs where variables are divided in two subsets which serve as inputs of an easily invertible (usually affine) triangular transformation. 
While such a transformation is invertible, its Jacobian is very sparse and thus may lack expressiveness.
This work presents a simple remedy by noting that subdivision and (affine) coupling 
can be repeated recursively within the resulting subsets, leading to an efficiently invertible block with dense, triangular Jacobian.
By formulating our recursive coupling scheme via a hierarchical architecture, HINT allows sampling from a joint distribution $p(\y,\x)$ and the corresponding posterior $p(\x\,|\,\y)$ using a single invertible network.
We evaluate our method on some standard data sets and benchmark its full power for density estimation and Bayesian inference on a novel data set of 2D shapes in Fourier parameterization, which enables consistent visualization of samples for different dimensionalities.
\end{abstract}

\section{Introduction}

Invertible neural networks based on the normalizing flow principle have recently gained increasing attention
for generative modeling,
in particular networks 
built on a coupling block design \cite{dinh2017density}.
Their success is due to a number of useful properties:
%
\rebuttal{\emph{(a)} they can tractably model complex high-dimensional probability densities without suffering from the curse-of-dimensionality,}
\emph{(b)} training via the maximum likelihood objective is generally very stable,
\emph{(c)} their latent space opens up opportunities for model interpretation and manipulation, and
\emph{(d)} the same trained model can be used for both efficient data generation and efficient density calculation.

While autoregressive models can also be trained as normalizing flows and share properties \emph{(a)} and \emph{(b)},
they sacrifice efficient invertibility for expressive power and thus lose properties \emph{(c)} and \emph{(d)}.
In contrast, lack of expressive power of a single invertible block is a core limitation of invertible networks, which needs to be compensated by extremely deep models with dozens or hundreds of blocks, e.g.,
the GLOW architecture \cite{kingma2018glow}.
While invertibility allows to back-propagate through very deep networks with minimal memory footprint \cite{gomez2017reversible}, more expressive invertible building blocks are still of great interest.
The superior performance of autoregressive approaches such as \cite{van2016conditional} is due to the stronger interaction 
between variables, reflected in a dense triangular Jacobian matrix,
\rebuttal{at the expense of cheap inversion}. 
The theory of transport maps \cite{villani2008optimal} provides certain guarantees of universality for triangular maps, which
do not hold for the standard coupling block design with a comparatively sparse Jacobian (\cref{fig:jacobians}, \emph{left}).

\begin{figure}[t]
    \centering
    \begin{subfigure}{0.35\linewidth}
        \resizebox{\textwidth}{!}{\begin{tikzpicture}[
    every node/.style = {inner sep=0pt, outer sep=0pt, anchor=center, align=center, font=\sffamily\scriptsize, text=black!67},
    box/.style = {rectangle, minimum width=1cm, minimum height=1cm, draw=white, line width=1pt}]


    \node [box, fill=black!20] (12) at (1,-2) {};

    \foreach \i in {0.0, 0.03125, ..., 1.96875}
        \node [box, fill=black!50, minimum height=0.3125mm, minimum width=0.3125mm, line width=0pt]
            at (0.515625+\i, -0.515625-\i) {};

    \draw [black, line width=1pt] (0.5, -0.5) rectangle (2.5, -2.5);

    \node [anchor=north] at (1.5,-2.6) {$\x$};
    \node [anchor=south, rotate=90] at (0.4,-1.5) {$\nabla_\bullet\,f_\text{C}(\x)$};
    
\end{tikzpicture} }%
    \end{subfigure}%
    \hspace{.03\linewidth}
    \begin{subfigure}{0.35\linewidth}
        \resizebox{\textwidth}{!}{\begin{tikzpicture}[
    every node/.style = {inner sep=0pt, outer sep=0pt, anchor=center, align=center, font=\sffamily\scriptsize, text=black!67},
    box/.style = {rectangle, minimum width=1cm, minimum height=1cm, draw=white, line width=.4pt}]

    \draw [black!50, line width=1pt] (0.5, -0.5) -- (2.5, -2.5);

    \node [box, fill=black!20] (12) at (1,-2) {};

    \node [box, fill=black!30, minimum height=5mm, minimum width=5mm] at (.75,-1.25) {};
    \node [box, fill=black!30, minimum height=5mm, minimum width=5mm] at (1.75,-2.25) {};

    \foreach \i in {0.0, 0.5, 1.0, 1.5}
        \node [box, fill=black!35, minimum height=2.5mm, minimum width=2.5mm]
            at (.625+\i, -.875-\i) {};

    \foreach \i in {0.0, 0.25, ..., 1.75}
        \node [box, fill=black!40, minimum height=1.25mm, minimum width=1.25mm]
            at (.5625+\i, -.6875-\i) {};

    \foreach \i in {0.0, 0.125, ..., 1.875}
        \node [box, fill=black!45, minimum height=0.625mm, minimum width=0.625mm]
            at (0.53125+\i, -0.59375-\i) {};

    \foreach \i in {0.0, 0.0625, ..., 1.9375}
        \node [box, fill=black!50, minimum height=0.3125mm, minimum width=0.3125mm]
            at (0.515625+\i, -0.546875-\i) {};

    \draw [black, line width=1pt] (0.5, -0.5) rectangle (2.5, -2.5);

    \node [anchor=north] at (1.5,-2.6) {$\x$};
    \node [anchor=south, rotate=90] at (0.4,-1.5) {$\nabla_\bullet\,f_\text{R}(\x)$};

\end{tikzpicture} }%
    \end{subfigure}%
    \caption{Sparse \emph{(left)} and dense \emph{(right)} triangular Jacobian of a standard coupling block and of our recursive design, respectively. Nonzero parts of the Jacobian in gray.}
    \label{fig:jacobians}
\end{figure}

Here, we propose an extension to the coupling block design that recursively fills in the previously unused portions of the Jacobian using smaller coupling blocks.
This allows for dense triangular maps (\cref{fig:jacobians}, \emph{right}),
or any intermediate design if the recursion is stopped before, while retaining
the advantages of the original coupling block architecture.
Furthermore, the recursive structure of this mapping can be used for efficient conditional sampling and Bayesian inference. Splitting the variables of interest into two subsets $\x$ and $\y$, a single normalizing flow model can be built that allows efficient sampling from both the joint distribution $p(\x,\y)$ and the conditional $p(\x\,|\,\y)$.
\rebuttal{It should be noted that our extension would also work for convolutional architectures like GLOW.}

Finally, we introduce a new family of data sets based on Fourier parameterizations of two-dimensional curves.
In the normalizing flow literature, there is an abundance of two-dimensional toy densities that provide an easy visual check for correctness of the model output.
However, the sparsity of the basic coupling block only becomes an issue beyond two dimensions where it is challenging to visualize the distribution or individual samples.
Pixel-based image data sets, on the other hand, quickly are too high dimensional for a meaningful assessment of the quality of the estimated densities.

A step towards visualizable data sets of intermediate size has been made in \cite{kruse2019benchmarking},
but their four-dimensional problems are still too simple to demonstrate the advantages of the recursive coupling approach described above.
To fill the gap, we describe a way to generate data sets of arbitrary dimension, where each data point parameterizes a closed curve in 2D space that is easy to visualize.
Increasing the input data dimension allows the representation of distributions of more and more complex curves.

To summarize, the contributions of this paper are: 
    \emph{(a)} a simple, efficiently invertible flow model with dense, triangular Jacobian;
    \emph{(b)} a hierarchical architecture to model joint as well as conditional distributions;
    \emph{(c)} a novel family of data sets allowing easy visualization for arbitrary dimensions.

The remainder of this work consists of a literature review, some mathematical background, a description of our method and supporting numerical experiments, followed by closing remarks.

\section{Related Work}
\label{sec:related-work}

Normalizing flows were popularized in the context of deep learning chiefly by the work of \cite{rezende2015variational} and \cite{dinh2015nice}.
By now, a large variety of architectures exist to realize normalizing flows. 
The majority 
falls into one of two groups: coupling block architectures and autoregressive models.
For a comprehensive overview and background information on invertible neural networks and normalizing flows see \cite{kobyzev2019normalizing} or \cite{papamakarios2019normalizing}.

Additive and then affine coupling blocks were first introduced by \cite{dinh2015nice,dinh2017density}, while \cite{kingma2018glow} went on to generalize the permutation of variables between blocks by learning the corresponding matrices, besides demonstrating the power of flow networks as generators.
Subsequent works have focused on replacing the (componentwise) affine transformation at the heart 
of such networks, which limits expressiveness, e.g., by replacing affine couplings with more expressive 
monotonous splines \cite{durkan2019neural}, albeit at the cost of evaluation speed.

On the other hand, there is lso a rich body of work on autoregressive (flow) networks
\cite{huang2018neural,kingma2016improved,van2016conditional,van2016pixel,papamakarios2017masked}.
More recently, \cite{jaini2019sum} applied second-order polynomials to improve expressive power over typical autoregressive models and proved that their model is a universal density approximator.
While such models provide excellent density estimation compared to coupling architectures \cite{liao2019generative,ma2019macow}, generating samples is often not a priority and can be prohibitively slow.

There are other approaches, outside those two subfields, that also seek a favorable trade-off between expressive power and efficient invertibility.
Residual Flows \cite{behrmann2019invertible,chen2019residual} impose Lipschitz constraints on a standard residual block, which guarantees invertibility with a full Jacobian and enables approximate maximum-likelihood training but requires an iterative procedure for sampling.
Similarly, \cite{song2019mintnet} uses lower triangular weight matrices that can be inverted via fixed-point iteration. The normalizing flow principle is formulated continuously as a differential equation (DE) by \cite{grathwohl2018scalable}, which allows free-form Jacobians but requires integrating a DE for each network pass.
\cite{karami2019invertible} introduce another method with dense Jacobian, based on invertible convolutions in the Fourier domain.

In terms of modeling conditional densities with invertible neural networks,
\cite{ardizzone2018analyzing} proposed an approach that divides the network output into conditioning variables and a latent vector, training the flow part with a \emph{maximum mean discrepancy} objective \citep[\textsc{MMD},][]{gretton2012kernel} instead of maximum likelihood.
Later \cite{ardizzone2019guided} introduced a simple conditional coupling block to construct a conditional normalizing flow.

\section{Mathematical Background}
\label{sec:background}

For an input vector $\x \in \mathbb{R}^{N}$, a standard, invertible coupling block is abstractly defined by
\begin{linenomath*}\begin{equation}
    \x' = f_\text{C}(\x) = \begin{bmatrix} \x_1 \\ C \big( \x_2 \,|\, \x_1 \big) \end{bmatrix} = \begin{bmatrix} \x'_1 \\ \x'_2 \end{bmatrix},
    \label{eq:coupling-block}
\end{equation}\end{linenomath*}
where $\x_1 = \x_{0:\lfloor N/2 \rfloor}$ and $\x_2 = \x_{\lfloor N/2 \rfloor:N}$ are the first and second half of the input vector and $\x'_2 = C(\x_2 \,|\, \x_1)$ is an easily invertible transform of $\x_2$ conditioned on $\x_1$.
Its inverse is then simply given by
\begin{linenomath*}\begin{equation}
    \x = f_\text{C}^{\sminus 1}(\x') = \begin{bmatrix} \x'_1 \\ C^{\sminus 1} \big( \x'_2 \,|\, \x'_1 \big) \end{bmatrix}.
    \label{eq:coupling-block-inverse}
\end{equation}\end{linenomath*}
For affine coupling blocks \cite{dinh2017density}, $C$ takes the form $C(\mathbf{u} \,|\, \mathbf{v}) = \mathbf{u} \odot \exp\big( s(\mathbf{v}) \big) + t(\mathbf{v})$ with $s$ and $t$ unconstrained feed-forward networks.
The logarithm of the Jacobian determinant of such a block can be computed very efficiently as
\begin{linenomath*}\begin{equation}
    \log \big| \det \mathbf{J}_{f_\text{C}}(\x) \big| = \log \left| \det \frac{\partial f_\text{C}(\x)}{\partial \x} \right| = \text{sum}( s\!\left( \x_1 \right) ).
    \label{eq:coupling-block-jacobian}
\end{equation}\end{linenomath*}
To ensure that all entries of $\x$ are transformed and interact with each other,
a pipeline that alternates between coupling blocks and random orthogonal matrices $\mathbf{Q}$ is constructed,
where the orthogonal block $\x' = f_\mathbf{Q}(\x) = \mathbf{Qx}$
can trivially be inverted as $\x = f^{\sminus 1}_\mathbf{Q}(\x') = \mathbf{Q}^\top \x'$ with log-determinant $\log \left| \det \mathbf{J}_{f_\mathbf{Q}}(\x) \right| = 0$.

\subsection{Normalizing Flows and Transport Maps}
\label{sec:norm-flow-transport}

To create a normalizing flow, this `pipeline' $T = f_{\text{C}1} \circ f_{\mathbf{Q}1} \circ f_{\text{C}2} \circ f_{\mathbf{Q}2} \circ \ldots$ is trained via maximum likelihood loss
\begin{linenomath*}\begin{equation}
    \mathcal{L}(\x) = \tfrac{1}{2} \|T(\x)\|_2^2 - \log \left| \mathbf{J}_T (\x) \right|
    \label{eq:max-likelihood-loss}
\end{equation}\end{linenomath*}
to transport the data distribution $p_X$ to a standard normal latent distribution $p_Z = \mathcal{N}(\mathbf{0}, \mathbf{I})$.
The map $T$ can then be used to sample from $p_X$ by drawing a sample $\z^{(i)}$ from $p_Z$ in the latent space and by passing it through the inverse model $S=T^{\sminus 1}$ to obtain $\x^{(i)} = S(\z^{(i)})$.

Using the change-of-variables formula, the density at a given data point $\x$ can also be calculated as $p_X(\x) = p_Z(T(\x)) \cdot \left| \det \mathbf{J}_T(\x) \right|$.
The mathematical basis of this procedure is the theory of transport maps \cite{villani2008optimal}, which are employed in exactly the same way to push a reference density (e.g.~Gaussian) to a target density (e.g.~the data distribution, \cite{marzouk2016sampling}). In fact, up to a constant, namely the (typically inaccessible) fixed entropy $H(p_X)$ of the data distribution, the expected value of the objective in \cref{eq:max-likelihood-loss} is the \emph{Kullback-Leibler (KL) divergence} between the data distribution $p_X$ and the push-forward of the latent density $S_\# p_Z$:
\begin{linenomath*}\begin{align}
    D_\text{KL}( p_X \,\|\, S_\# p_Z )
    &= \int p_X(\x) \log \frac{p_X(\x)}{S_\# p_Z(\x)} \,\text{d}\x \nonumber\\ 
    &= \mathbb{E}_{\x \sim p_X} [\mathcal{L}(\x)] + H(p_X). 
\end{align}\end{linenomath*}
Normalizing flows represent one parametrised family of maps over which \cref{eq:max-likelihood-loss} can be minimized. Other examples include polynomial \cite{marzouk2016sampling}, kernel-based \cite{liu2016stein} or low-rank tensor \cite{dolgov2020tt} approximations.

Note also that each pair $f_{\text{C}i} \circ f_{\mathbf{Q}i}$ in $T$ is a composition of an orthogonal transformation and a triangular map,
where the latter is better known in the field of transport maps as a \emph{Knothe-Rosenblatt rearrangement} \cite{marzouk2016sampling}.
This can be interpreted as a non-linear generalization of the classic QR decomposition \cite{stoer2013introduction}.
Whereas the triangular part encodes the possibility to represent non-linear transformations,
the orthogonal part reshuffles variables to foster dependence of each part of the input to the final output,
thereby drastically increasing the representational power of the map $T$.

\subsection{Bayesian Inference with Conditional Flows}
\label{sec:bayesian-setting}

Inverse problems arise when one possesses a well-understood model for the {\em forward mapping} $\x \rightarrow \y$ from hidden parameters $\x$ to observable outcomes $\y$, e.g.~in the form of an explicit likelihood $p(\y \,|\, \x)$ or a Monte-Carlo simulation.
However, the actual object of interest is the {\em inverse mapping} $\y \rightarrow \x$ from observations to parameters.
According to Bayes' theorem, this requires estimation of the posterior conditional density $p(\x \,|\, \y)$.
Such Bayesian inference problems arise frequently in the sciences and are generally very hard.

Normalizing flows can be used in several ways to estimate conditional densities.
The approach in this paper is inspired by \cite{marzouk2016sampling} and exploits the link to Knothe-Rosenblatt maps highlighted above.
As described below, see \cref{fig:hierarchical-coupling-block} (left),
it suffices to constrain the possible rearrangements of variables in the coupling blocks,
i.e. the choice of the orthogonal blocks $f_{\mathbf{Q}i}$, to enable conditional sampling.
This was first noted in \cite{detommaso2019hint1}. 

Independently, 
\cite{ardizzone2019guided} and \cite{winkler2019learning} introduced \emph{conditional} coupling blocks that allow an entire normalizing flow to be conditioned on external variables.
By conditioning the transport $T$ between $p_X(\x)$ and $p_Z(\z)$ on the corresponding values of $\y$ as $\z = T(\x \,|\, \y)$, its inverse $T^{\sminus 1}(\z \,|\, \y)$ can be used to turn the latent distribution $p_Z(\z)$ into an approximation of the posterior $p(\x \,|\, \y)$. 

\section{Method}
\label{sec:method}

We extend the basic coupling block design in two ways.

\subsection{The Recursive Coupling Block}

As visualized in \cref{fig:jacobians} \emph{(left)}, the Jacobian $\mathbf{J}_f$ of a simple coupling block is very sparse, i.e.~many possible interactions between variables are not modelled.
However the efficient determinant computation in \cref{eq:coupling-block-jacobian} works for any lower triangular $\mathbf{J}_f$,
and indeed theorem 1 of \cite{hyvarinen1999nonlinear} states that a single triangular transformation can, in theory, already represent arbitrary distributions.

The following recursive coupling scheme $f_{\text{R}}$ makes use of this potential and fills the empty areas below the diagonal: 
Given $\x \in \mathbb{R}^{N}$ and a hierarchy depth $K \in \mathbb{N}$, we define recursively, for $k=K, K-1, \ldots, 1$:
\begin{linenomath*}
    \begin{numcases}{\x' = f_{\text{R},k}(\x) =}
        f_\text{C}(\x), & if $N_k\!\leq\!3$, \nonumber\\
        \begin{bmatrix} f_{\text{R},k \sminus 1}( \x_1 ) \\ C_k \left( f_{\text{R},k \sminus 1} \big( \x_2 \big) \,\middle|\, \x_1 \right) \end{bmatrix}, & else,
    \label{eq:coupling-block-recursive}
    \end{numcases}
\end{linenomath*}
where for each $k$, $\x_1 = \x_{0:\lfloor N_k/2 \rfloor}$ and $\x_2 = \x_{\lfloor N_k/2 \rfloor:N_k}$, $N_k$ is the size of the current input vector and $N_K = N$.
Note that each sub-coupling has its own coupling function $C_k$ with independent parameters.
The inverse transform is
\begin{linenomath*}\begin{align}
    \x = f^{\sminus 1}_{\text{R},k}(\x') \!=\!
    \begin{cases}
        f^{\sminus 1}_\text{C}(\x'), \hfill \text{if } N_k \leq 3, \\
        \!\begin{bmatrix} f^{\sminus 1}_{\text{R},k \sminus 1} ( \x'_1 ) \\ f^{\sminus 1}_{\text{R},k \sminus 1} \!\left( C_k^{\sminus 1} \!\left( \x'_2 \,\middle|\, f^{\sminus 1}_{\text{R},k \sminus 1}(\x'_1) \right) \right) \end{bmatrix}\!, \text{ else}.
    \end{cases}
    \label{eq:coupling-block-recursive-inverse}
\end{align}\end{linenomath*}
For $K = \lceil \log_2 N \rceil$, this procedure leads to the dense lower triangular Jacobian visualized in \cref{fig:jacobians} \emph{(right)}, the log-determinant of which is simply the sum of the log-determinants of all sub-couplings $C_k$.
A visual representation of the architecture can be seen in \cref{fig:recursive-coupling}.

\begin{figure}[t]
    \centering
    \resizebox{0.85\linewidth}{!}{\begin{tikzpicture}[
    every node/.style={inner sep=0pt, outer sep=0pt, anchor=center, align=center, font=\sffamily\Huge, text=black!75},
    var/.style={rectangle, minimum width=2em, minimum height=2em, text depth=0, line width=1pt, draw=black!50, fill=black!5},
    op/.style={circle, minimum width=2em, text depth=2pt, line width=1pt, draw=black, fill=black!50, text=black!5, font=\huge\boldmath},
    nn/.style={op, regular polygon, regular polygon sides=6, inner sep=-4pt, minimum width=1.7cm, fill=vll-dark},
    dot/.style={circle, minimum width=5pt, fill=vll-dark},
    connect/.style={line width=1pt, draw=vll-dark},
    arrow/.style={connect, -{Triangle[length=6pt, width=5pt]}}]

    \node [anchor=east] (yin) at (0,0) {$\x$};

    \node [op, inner sep=4pt] (yperm) at ([shift={(yin.east)}] 0:1.2) {$\mathbf{Q}$};
    \draw [connect, shorten <=3pt] (yin) to (yperm);

    \node [dot] (ysplit) at ([shift={(yperm.east)}] 0:1.4) {};
    \draw [connect] (yperm) to (ysplit);

    \node (yu1) at ([shift={(ysplit)}] 60:1.5) {};
    \node (yu2) at ([shift={(ysplit)}] -60:1.5) {};
    \node [dot] (yd1) at ([shift={(yu1)}] 0:2.5) {};
    \node [dot] (yd2) at ([shift={(yd1)}] 0:2) {};

    \node [op] (ymult1) at ([shift={(yd1)}] -60:3) {$\odot$};
    \draw [arrow] (yd1) -- (ymult1);
    \draw [arrow] (ysplit) -- (yu2.center) -- (ymult1);

    \node [op] (yadd1) at ([shift={(yd2)}] -60:3) {$+$};
    \draw [arrow] (yd2) -- (yadd1);
    \draw [arrow] (ymult1) -- (yadd1);

    \node (yv1) at ([shift={(yadd1)}] 0:2.5) {};
    \node (yv2) at (yv1 |- yu1) {};
    \node [dot] (ycat) at ([shift={(yv1)}] 60:1.5) {};
    \draw [connect] (yadd1) -- (yv1.center) -- (ycat);
    \path [connect] (ysplit) -- (yu1.center) -- (yu1.center -| yv1.center) -- (ycat);

    \node [anchor=west] (yout) at ([shift={(ycat.east)}] 0:1.5) {$\mathbf{Q} \cdot f_\text{R}(\x)$};
    \draw [arrow, shorten >=3pt] (ycat) to (yout);

    \node [nn] (ys2) at ([shift={(yd1)}] -60:1.3) {\scalebox{0.6}{$\vphantom{t}s(\x_1)$}};
    \node [nn] (yt2) at ([shift={(yd2)}] -60:1.3) {\scalebox{0.6}{$t(\x_1)$}};

    \node [nn, scale=0.7] (T1) at ([shift={(yu1.center -| yv1.center)}] -180:1.5) {$f_\text{R}(\x_1)$};
    \node [nn, scale=0.7] (T2) at ([shift={(yu2.center)}] 0:1.5) {$f_\text{R}(\x_2)$};

    \begin{pgfonlayer}{background}
        \draw [draw=black!67, fill=black!10]
            ([shift={(ysplit)}] 180:1) --
            ([shift={(yu1)}] 90:1.732) --
            ([shift={(yv2)}] 90:1.732) --
            ([shift={(ycat)}] 0:1) --
            ([shift={(yv1)}] -90:1.732) --
            ([shift={(yu2)}] -90:1.732) --
            cycle;
        \node [anchor=east, black!33] at ([shift={(yv1)}] -120:1) {$f_\text{R}$};
    \end{pgfonlayer}


\end{tikzpicture} }%
    \caption{A recursive affine coupling block. The inner functions $f_\text{R}(\x_i)$ take again the form of the outer gray block, repeated until the maximum hierarchy depth is reached. 
    Each such coupling block in itself has a triangular Jacobian.}
    \label{fig:recursive-coupling}
\end{figure}

\begin{figure*}[t]
    \centering
       \resizebox{0.35\linewidth}{!}{\begin{tikzpicture}[
    every node/.style={inner sep=0pt, outer sep=0pt, anchor=center, align=center, font=\sffamily\huge, text=black!75},
    var/.style={rectangle, minimum width=2em, minimum height=2em, text depth=0, line width=1pt, draw=black!50, fill=black!5},
    op/.style={circle, minimum width=2em, text depth=2pt, line width=1pt, draw=black, fill=black!50, text=black!5, font=\huge\boldmath},
    nn/.style={op, regular polygon, regular polygon sides=6, inner sep=-2pt, minimum size=2.2cm, fill=vll-dark},
    dot/.style={circle, minimum width=5pt, fill=vll-dark},
    connect/.style={line width=1pt, draw=vll-dark},
    arrow/.style={connect, -{Triangle[length=6pt, width=4pt]}}]

    \node [anchor=east] (yin) at (0,0) {$\y$};

    \node [op, minimum width=3.5em] (yperm) at ([shift={(yin.east)}] 0:1.5) {$\scriptstyle\mathbf{Q}_\y$};
    \draw [connect, shorten <=3pt] (yin) to (yperm);

    \node [anchor=west] (yout) at ([shift={(yperm.east)}] 0:7.7) {$\mathbf{Q}_\y \cdot f_{\text{H},\y}(\y)$};
    \draw [arrow, shorten >=3pt] (yperm) to (yout);

    \node [nn, scale=0.7, fill=vll-orange] (ynn) at ([shift={(yout.west)}] 180:1.8) {$f_\text{R}(\y)$};

    \node [dot] (con1) at ([shift={(yperm)}] 0:1.9) {};
    \node [dot] (con2) at ([shift={(yperm)}] 0:4) {};
    \node [op] (mult) at ([shift={(con1)}] -60:5) {$\odot$};
    \node [op] (add) at ([shift={(con2)}] -60:5) {$+$};
    \draw [arrow] (con1) -- (mult);
    \draw [arrow] (con2) -- (add);

    \node [nn, scale=0.9, fill=mpl-blue] (scon) at ([shift={(mult)}] 120:2.6) {$\bm{s}(\y)$};
    \node [nn, scale=0.9, fill=mpl-blue] (tcon) at ([shift={(add)}] 120:2.6) {$\bm{t}(\y)$};

    \node [anchor=east] (xin) at (mult -| yin.east) {$\x$};

    \node [op, minimum width=3.5em] (xperm) at ([shift={(xin.east)}] 0:1.5) {$\scriptstyle\mathbf{Q}_\x$};
    \draw [connect, shorten <=3pt] (xin) to (xperm);

    \node [anchor=west] (xout) at (xin -| yout.west) {$\mathbf{Q}_\x \cdot f_{\text{H},\x}(\x \,|\, \y)$};
    \begin{scope}[on background layer]
        \draw [arrow, shorten >=3pt] (xperm) to (xout);
    \end{scope}

    \node [nn, scale=0.7, fill=vll-green] (xnn) at ([shift={(xperm.east)}] 0:1.5) {$f_\text{R}(\x)$};

\end{tikzpicture} }%
       \hspace{0.05\linewidth}%
       \resizebox{0.15\linewidth}{!}{\begin{tikzpicture}[
    every node/.style = {inner sep=0pt, outer sep=0pt, anchor=center, align=center, font=\sffamily\scriptsize, text=black!67},
    box/.style = {rectangle, draw=white, line width=.4pt, anchor=north east}]

    \draw [vll-light, line width=.5pt, dotted] (.5, -.5) rectangle (2, -.5);
    \draw [vll-light, line width=.5pt, dotted] (.5, .0) rectangle (.5, -.5);

    \draw [vll-orange!50!black, line width=1pt] (0, 0) -- (.5, -.5);
    \draw [vll-green!50!black, line width=1pt] (.5, -.5) -- (2, -2);

    \node [box, fill=mpl-blue, minimum height=15mm, minimum width=5mm] (12) at (.5,-.5) {};

    \node [box, fill=vll-orange, minimum height=2.5mm, minimum width=2.5mm] at (.25,-.25) {};
    \node [box, fill=vll-green, minimum height=7.5mm, minimum width=7.5mm] at (1.25,-1.25) {};

    \foreach \i in {.0, .25}
        \node [box, fill=vll-orange!85!black, minimum height=1.25mm, minimum width=1.25mm]
            at (.125+\i, -.125-\i) {};
    \foreach \i in {0.0, 0.75}
        \node [box, fill=vll-green!85!black, minimum height=3.75mm, minimum width=3.75mm]
            at (.875+\i, -.875-\i) {};

    \foreach \i in {.0, .125, .25, .375}
        \node [box, fill=vll-orange!70!black, minimum height=.625mm, minimum width=0.625mm]
            at (.0625+\i, -.0625-\i) {};
    \foreach \i in {.0, .375, .75, 1.125}
        \node [box, fill=vll-green!70!black, minimum height=1.875mm, minimum width=1.875mm]
            at (.6875+\i, -.6875-\i) {};

    \foreach \i in {.0, .0625, ..., .49}
        \node [box, fill=vll-orange!60!black, minimum height=0.3125mm, minimum width=0.3125mm]
            at (.03125+\i, -.03125-\i) {};
    \foreach \i in {.0, .1875, ..., 1.49}
        \node [box, fill=vll-green!60!black, minimum height=.9375mm, minimum width=.9375mm]
            at (.59375+\i, -.59375-\i) {};

    \foreach \i in {.0, .09375, ..., 1.49}
        \node [box, fill=vll-green!50!black, minimum height=.46875mm, minimum width=.46875mm]
            at (.546875+\i, -.546875-\i) {};

    \foreach \i in {.0, .046875, ..., 1.49}
        \node [box, fill=vll-green!50!black, minimum height=.234375mm, minimum width=.234375mm]
            at (.5234375+\i, -.5234375-\i) {};

    \draw [black, line width=1pt] (0, 0) rectangle (2, -2);

    \node [anchor=north] at (.25, -2.1) {$\y$};
    \node [anchor=north] at (1.25, -2.1) {$\x$};
    \node [anchor=south, rotate=90, scale=.6] at (-.1, -.25) {$\nabla_\bullet\,f_\text{H}(\y)$};
    \node [anchor=south, rotate=90, scale=.6] at (-.1, -1.25) {$\nabla_\bullet\,f_\text{H}(\x)$};

\end{tikzpicture} } \\[7pt]
       \resizebox{.9\linewidth}{!}{\begin{tikzpicture}[
    every node/.style={inner sep=0pt, outer sep=0pt, text depth=0pt, anchor=center, align=center, font=\sffamily\huge, text=black!75},
    var/.style={rectangle, minimum width=2em, minimum height=2em, text depth=0, line width=1pt, draw=black!50, fill=black!5},
    op/.style={circle, minimum width=2em, text depth=2pt, line width=1pt, draw=black, fill=black!50, text=black!5, font=\huge\boldmath},
    nn/.style={op, regular polygon, regular polygon sides=6, inner sep=-2pt, minimum size=1.8cm, fill=vll-dark},
    dot/.style={circle, minimum width=5pt, fill=vll-dark},
    connect/.style={line width=1pt, draw=vll-dark},
    arrow/.style={connect, -{Triangle[length=6pt, width=4pt]}}]

    \node [anchor=east] (yin) at (0,0) {$\y$};
    \node [anchor=west] (yout) at ([shift={(yin.east)}] 0:19.5) {$\z_\y$};
    \begin{scope}[on background layer]
        \draw [arrow, shorten <=3pt, shorten >=3pt] (yin) to (yout);
    \end{scope}

    \node [op, minimum width=3.5em] (yperm1) at ([shift={(yin.east)}] 0:1.5) {};
    \node [dot] (con1) at ([shift={(yperm1)}] 0:1.9) {};
    \node [op] (op1) at ([shift={(con1)}] -60:4) {};
    \node [nn, fill=vll-orange] (ynn1) at ([shift={(yin.east)}] 0:5.5) {};

    \node [op, minimum width=3.5em] (yperm2) at ([shift={(yin.east)}] 0:7.5) {};
    \node [dot] (con2) at ([shift={(yperm2)}] 0:1.9) {};
    \node [op] (op2) at ([shift={(con2)}] -60:4) {};
    \node [nn, fill=vll-orange] (ynn2) at ([shift={(yin.east)}] 0:11.5) {};

    \node [op, minimum width=3.5em] (yperm3) at ([shift={(yin.east)}] 0:13.5) {};
    \node [dot] (con3) at ([shift={(yperm3)}] 0:1.9) {};
    \node [op] (op3) at ([shift={(con3)}] -60:4) {};
    \node [nn, fill=vll-orange] (ynn3) at ([shift={(yin.east)}] 0:17.5) {};

    \draw [connect, dotted] (con1) -- (op1);
    \node [nn] (connn1) at ([shift={(op1)}] 120:2.1) {};

    \draw [connect, dotted] (con2) -- (op2);
    \node [nn] (connn2) at ([shift={(op2)}] 120:2.1) {};

    \draw [connect, dotted] (con3) -- (op3);
    \node [nn] (connn3) at ([shift={(op3)}] 120:2.1) {};

    \node [anchor=east] (xin) at (op1 -| yin.east) {};
    \node [anchor=west] (xout) at (xin -| yout.west) {};
    \begin{scope}[on background layer]
        \draw [connect, shorten <=3pt, shorten >=3pt, dotted] (xin) to (xout);
    \end{scope}

    \node [op, minimum width=3.5em] (xperm1) at ([shift={(xin.east)}] 0:1.5) {};
    \node [nn] (xnn1) at ([shift={(xperm1.east)}] 0:1.5) {};

    \node [op, minimum width=3.5em] (xperm2) at ([shift={(xin.east)}] 0:7.5) {};
    \node [nn] (xnn1) at ([shift={(xperm2.east)}] 0:1.5) {};

    \node [op, minimum width=3.5em] (xperm2) at ([shift={(xin.east)}] 0:13.5) {};
    \node [nn] (xnn2) at ([shift={(xperm2.east)}] 0:1.5) {};

\end{tikzpicture} \hspace{3cm} \begin{tikzpicture}[
    every node/.style={inner sep=0pt, outer sep=0pt, text depth=0pt, anchor=center, align=center, font=\sffamily\huge, text=black!75},
    var/.style={rectangle, minimum width=2em, minimum height=2em, text depth=0, line width=1pt, draw=black!50, fill=black!5},
    op/.style={circle, minimum width=2em, text depth=2pt, line width=1pt, draw=black, fill=black!50, text=black!5, font=\huge\boldmath},
    nn/.style={op, regular polygon, regular polygon sides=6, inner sep=-2pt, minimum size=1.8cm, fill=vll-dark},
    dot/.style={circle, minimum width=5pt, fill=vll-dark},
    connect/.style={line width=1pt, draw=vll-dark},
    arrow/.style={connect, -{Triangle[length=6pt, width=4pt]}}]

    \node [anchor=east] (yin) at (0,0) {};
    \node [anchor=west] (yout) at ([shift={(yin.east)}] 0:19.5) {$\z_\y$};

    \node [op, minimum width=3.5em] (yperm1) at ([shift={(yin.east)}] 0:1.5) {};
    \node [dot] (con1) at ([shift={(yperm1)}] 0:1.9) {};
    \node [op] (op1) at ([shift={(con1)}] -60:4) {};
    \node [nn] (ynn1) at ([shift={(yin.east)}] 0:5.5) {};

    \node [op, minimum width=3.5em] (yperm2) at ([shift={(yin.east)}] 0:7.5) {};
    \node [dot] (con2) at ([shift={(yperm2)}] 0:1.9) {};
    \node [op] (op2) at ([shift={(con2)}] -60:4) {};
    \node [nn] (ynn2) at ([shift={(yin.east)}] 0:11.5) {};

    \node [op, minimum width=3.5em] (yperm3) at ([shift={(yin.east)}] 0:13.5) {};
    \node [dot] (con3) at ([shift={(yperm3)}] 0:1.9) {};
    \node [op] (op3) at ([shift={(con3)}] -60:4) {};
    \node [nn] (ynn3) at ([shift={(yin.east)}] 0:17.5) {};

    \begin{scope}[on background layer]
        \draw [arrow, shorten <=3pt] (yout) to (con1);
        \draw [connect, shorten >=3pt, dotted] (con1) to (yin);
    \end{scope}

    \draw [arrow] (con1) -- (op1);
    \node [nn, fill=mpl-blue] (connn1) at ([shift={(op1)}] 120:2.1) {};

    \draw [arrow] (con2) -- (op2);
    \node [nn, fill=mpl-blue] (connn2) at ([shift={(op2)}] 120:2.1) {};

    \draw [arrow] (con3) -- (op3);
    \node [nn, fill=mpl-blue] (connn3) at ([shift={(op3)}] 120:2.1) {};

    \node [anchor=east] (xin) at (op1 -| yin.east) {$\x$};
    \node [anchor=west] (xout) at (xin -| yout.west) {\raisebox{5pt}{$\z_\x \sim \mathcal{N}(\z_\x; \mathbf{0}, \mathbf{I}_{|\x|})$}};
    \begin{scope}[on background layer]
        \draw [arrow, shorten <=3pt, shorten >=3pt] (xout) to (xin);
    \end{scope}

    \node [op, minimum width=3.5em] (xperm1) at ([shift={(xin.east)}] 0:1.5) {};
    \node [nn, fill=vll-green] (xnn1) at ([shift={(xperm1.east)}] 0:1.5) {};

    \node [op, minimum width=3.5em] (xperm2) at ([shift={(xin.east)}] 0:7.5) {};
    \node [nn, fill=vll-green] (xnn1) at ([shift={(xperm2.east)}] 0:1.5) {};

    \node [op, minimum width=3.5em] (xperm2) at ([shift={(xin.east)}] 0:13.5) {};
    \node [nn, fill=vll-green] (xnn2) at ([shift={(xperm2.east)}] 0:1.5) {};

\end{tikzpicture}}
    \caption{\emph{Top:} Single HINT block with recursive coupling, and its Jacobian matrix.
    Transformation of $\x$ is influenced by $\y$, but not vice-versa, imposing a hierarchy on variables.
    \emph{Bottom:} Using HINT flows for conditional sampling/Bayesian inference.}
    \label{fig:hierarchical-coupling-block}
\end{figure*}

However, since the (sub-)coupling blocks are affine, $f_{\text{R},K}$ can only represent an approximation of the exact Knothe-Rosenblatt map and it is still necessary,
as for standard coupling blocks,
to create a normalizing flow by composing several recursive coupling blocks interspersed with orthogonal transformations $f_\mathbf{Q}$.
Thus, in practice, it is also more economical to limit the depth of the hierarchy to~2 or~3.
This already increases the amount of interaction between individual variables considerably, while limiting the computational overhead, but it allows to use much shallower networks.
The trade-off between number of blocks and hierarchy depth will be studied for the Fourier shapes data set in this work.

\subsection{Hierarchical Invertible Neural Transport}

While the recursive coupling block defined above is motivated by the search for a more expressive architecture, it is also ideally suited for estimating conditional flows and thus for Bayesian inference.

Specifically, in a setting with paired data $(\x_i, \y_i)$, where subsequently we want a sampler for $\x$ conditioned on $\y$, we can provide both variables as input to the flow,
separating them in the first hierarchy level for further transformation at the next recursion level.
Crucially, $\x$ and $\y$ variables are never permuted between lanes, thus only feeding forward information from the $\y$-lane to the $\x$-lane as shown in \cref{fig:hierarchical-coupling-block} \emph{(top left)}.
Instead of one large permutation operation over all variables, as in the hierarchical coupling block design in \cref{fig:recursive-coupling},
we apply individual permutations $\mathbf{Q}_\y$ and $\mathbf{Q}_\x$ to each respective lane at the beginning of the block.
%
%
A normalizing flow model constructed in this way performs \emph{hierarchical invertible neural transport}, or HINT for short.

The output of a HINT model is a latent code with two components, $\z = [\z_\y, \z_\x]^\top = T(\y, \x)$,
but the training objective stays the same as in \cref{eq:max-likelihood-loss}:
\begin{linenomath*}\begin{equation}
    \mathcal{L}(\y,\x) = \tfrac{1}{2} \|T(\y,\x)\|_2^2 - \log \left| \mathbf{J}_T (\y,\x) \right|
    \label{eq:max-likelihood-loss-bayesian}
\end{equation}\end{linenomath*}
As with a standard normalizing flow, the joint density of input variables is the pull-back of the latent density via $T$:
\begin{linenomath*}\begin{equation}
    p_T(\y,\x) = S_\# p_Z(\z) =
    S_\# \mathcal{N}(\mathbf{0}, \mathbf{I}_{|\y|+|\x|}),
    \label{eq:hint-joint-density}
\end{equation}\end{linenomath*}
where $S=T^{-1}$. But because the $\y$-lane in HINT can be evaluated independently of the $\x$-lane,
we can determine the partial latent code $\z_\y$ for a given $\y$ and hold it fixed (\cref{fig:hierarchical-coupling-block}, \emph{bottom left}),
while drawing $\z_\x$ from the $\x$-part of the latent distribution \emph{(bottom right)}.
This yields samples from the conditional density:
\begin{linenomath*}\begin{equation}
    \x = S^{\x}([\z_\y, \z_x]) \sim p_T(\x \,|\, \y) \quad \text{with } \z_\y = T^\y(\y),
\end{equation}\end{linenomath*}
where superscripts $\x$ and $\y$ respectively denote $\x$- and $\y$-lanes of the transformations.
This means HINT gives access to both the joint density of $\x$ and $\y$, as well as the conditional density of $\x$ given $\y$, e.g.~for Bayesian inference.

\subsection{Computational Complexity}

\rebuttal{The number of couplings doubles in every recursion level, whereas the workload per coupling decreases exponentially, so that the total order of complexity of HINT and RealNVP is the same.
All sub-networks $s$ and $t$ within one level are independent of each other and can be processed in a single parallel pass on the GPU (see appendix).
Only the final affine transformations and some bookkeeping operations must be executed sequentially, but at negligible cost compared to the other tensor operations.
Our first, non-parallel implementation of HINT is 2-10 times slower than RealNVP.%
}

\section{Experiments}
\label{sec:experiments}

We perform experiments on classical UCI data sets \cite{dua2019UCI} and a new data set called ``Fourier shapes''.
We introduce this new data set to balance four conflicting goals:
\begin{enumerate}
    \item The dimension of the data should be high enough for HINT's hierarchical decomposition to make a difference.
    \item The dimension should be low enough to allow for accurate quantitative evaluation and comparison of results.
    \item Learning the joint distribution should be challenging due to complex interactions between variables.
    \item Visualizations should allow intuitive qualitative comparison
    of the differences between alternative approaches.
\end{enumerate}
Our new data set represents families of 2-dimensional contours in terms of the probability density of their Fourier coefficients and fulfills the above requirements: 
The dimension of the problem can be easily adjusted by controlling the complexity of the shapes under consideration and the number of Fourier coefficients (1,2).
Shapes with sharp corners and long-range symmetries require accurate alignment of many Fourier coefficients (specifically, of their phases, 3 -- see appendix).
Humans can readily recognize the quality of a shape representation in a picture (4).

\rebuttal{Our experiments show considerable improvements of HINT over RealNVP.
To clearly demonstrate these advantages, we heavily restrict the networks' parameter budgets -- larger networks would be much more accurate, but exhibit less meaningful differences.}
Models were trained on an \emph{RTX 2080 Ti} GPU.
Hyper-parameters are listed in the appendix.%
\footnote{\rebuttal{Code and data at \url{https://github.com/VLL-HD/HINT}.}}

\subsection{UCI Density Estimation Benchmarks}

The tabular UCI data sets \cite{dua2019UCI} are popular for comparing density models.
Using public code for pre-processing\footnote{https://github.com/LukasRinder/normalizing-flows}, we compare several flow models with \textsc{Real-NVP} and \textsc{recursive} coupling blocks in terms of the average log-likelihood on the test set.
To tease out shortcomings, each model is ``handicapped'' to a budget of 500k (\textsc{Power}, \textsc{Gas}) or 250k (\textsc{Miniboone}) trainable parameters.

\Cref{tab:density-estimation-uci} shows that the recursive design achieves similar or better test likelihood in all cases, even when the low dimensionality (\textsc{Dim}) of the data allows little recursion.

\begin{table}[t]
    \begin{center}\begin{small}\begin{sc}
    \begin{tabular}{l l@{~}
        c@{~}
        S[detect-weight,mode=text,table-format=-2.3(4)]@{\quad}
        S[detect-weight,mode=text,table-format=-2.3(4)]}
    \toprule
    Blocks\hspace{-0.75cm} 
    & & Dim~
    & {Real-NVP} & {Recursive} \\
    \midrule
    & Power 
    &  6 &  -0.054 \pm 0.017 & \B  -0.027 \pm 0.018 \\
    4 & Gas 
    &  8 &   7.620 \pm 0.136 & \B   7.662 \pm 0.094 \\
    & Miniboone 
    & 42 & -19.296
    \pm 0.395 & \B -14.547 
    \pm 0.164 \\
    \midrule
    & Power 
    &  6 &   \B 0.093 \pm 0.002 &   0.080 \pm 0.007 \\
    8 & Gas 
    &  8 &   8.062 \pm 0.177 &   \B 8.137 \pm 0.055 \\
    & Miniboone 
    & 42 & -16.625 
    \pm 0.119 & \B -14.117
    \pm 0.163 \\
    \bottomrule
    \end{tabular}
    \end{sc}\end{small}\end{center}
    \caption{Normal and recursive coupling compared on UCI benchmarks in terms of average log-likelihood (mean $\pm$ std over 3 training runs; higher is better $\uparrow$).}
    \label{tab:density-estimation-uci}
\end{table}

\subsection{Fourier Shapes Data Set}

A curve $\mathbf{g}(t) \in \mathbb{R}^2$, parameterized by $2M\!+\!1$ complex 2d Fourier coefficients $\mathbf{a}_m \in \mathbb{C}^2$, can be traced as
\begin{linenomath*}
\begin{equation}
    \mathbf{g}(t) = \textstyle\sum_{m = -M}^{M} \mathbf{a}_m \cdot e^{2\pi \cdot i \cdot m \cdot t}
    \label{eq:fourier-curve}
\end{equation}
\end{linenomath*}
with parameter $t$ running from $0$ to $1$. This parameterization will always yield a closed, possibly self-intersecting curve \cite{mcgarva1993harmonic}.

Vice-versa, we can calculate the Fourier coefficients
\begin{linenomath*}
\begin{equation}
    \mathbf{a}_m = \frac{1}{L} \sum_{l=0}^{L-1} \mathbf{p}_l \cdot e^{-2\pi \cdot i \cdot m \cdot l / L} , \  \text{for} \; m \in [-M, M],
    \label{eq:fourier-coefficients}
\end{equation}
\end{linenomath*}
to approximately fit a curve through a sequence of $L$ points $\mathbf{p}_l \in \mathbb{R}^2$, $l=0,\ldots, L-1$.
By increasing $M$, higher order terms are added to the parameterization in \cref{eq:fourier-curve} and the shape is approximated in greater detail.
An example of this effect for a natural shape is shown in \cref{fig:fourier-shape} \emph{(right)}.
Note that the actual dimensionality of the parameterization in our data set is $|\x| = 4 \cdot (2M+1)$, as each complex 2d coefficient $\mathbf{a}_m$ is represented by four real numbers.

\begin{figure}[t]
    \centering
    \begin{subfigure}[b]{0.3\linewidth}
        \resizebox{\textwidth}{!}{\begin{tikzpicture}[
    every node/.style={inner sep=0pt, outer sep=0pt, text depth=0pt, anchor=center, align=center, font=\sffamily\huge, text=black!75}]

    \filldraw [draw=vll-dark, fill=vll-light!10, line width=.2pt]
    (0.023923, 0.50239) -- (0.052632, 0.54067) -- (0.08134, 0.57416) -- (0.10526, 0.61244) -- (0.11962, 0.65072) -- (0.14833, 0.689) -- (0.1866, 0.70335) -- (0.22488, 0.69378) -- (0.23923, 0.6555) -- (0.27751, 0.67464) -- (0.311, 0.70335) -- (0.34928, 0.73684) -- (0.38756, 0.75598) -- (0.42584, 0.77512) -- (0.46411, 0.78947) -- (0.49761, 0.82775) -- (0.5311, 0.86603) -- (0.56938, 0.89474) -- (0.5933, 0.93301) -- (0.59809, 0.97129) -- (0.62679, 1.0) -- (0.66507, 1.0) -- (0.70335, 0.98565) -- (0.73684, 0.96651) -- (0.77512, 0.96172) -- (0.8134, 0.96172) -- (0.85167, 0.95694) -- (0.88517, 0.93301) -- (0.90431, 0.89474) -- (0.90431, 0.85646) -- (0.89952, 0.81818) -- (0.88995, 0.7799) -- (0.88995, 0.74163) -- (0.88517, 0.70335) -- (0.86603, 0.66507) -- (0.84689, 0.62679) -- (0.82297, 0.58852) -- (0.79904, 0.55502) -- (0.76077, 0.52632) -- (0.74641, 0.49282) -- (0.73684, 0.45455) -- (0.73684, 0.41627) -- (0.73684, 0.37799) -- (0.76555, 0.33971) -- (0.78469, 0.30622) -- (0.81818, 0.33971) -- (0.84211, 0.32057) -- (0.8756, 0.32536) -- (0.91388, 0.33014) -- (0.91866, 0.29665) -- (0.88038, 0.28708) -- (0.8756, 0.26316) -- (0.91388, 0.26316) -- (0.9378, 0.23923) -- (0.90431, 0.22967) -- (0.86603, 0.2201) -- (0.82775, 0.19139) -- (0.78947, 0.1866) -- (0.7512, 0.1866) -- (0.7177, 0.21053) -- (0.69378, 0.24402) -- (0.66507, 0.27273) -- (0.64115, 0.311) -- (0.61722, 0.31579) -- (0.58373, 0.2823) -- (0.54545, 0.25837) -- (0.50718, 0.24402) -- (0.50239, 0.21531) -- (0.52632, 0.19139) -- (0.56459, 0.1866) -- (0.57416, 0.15311) -- (0.61244, 0.13397) -- (0.63636, 0.10048) -- (0.61244, 0.08134) -- (0.65072, 0.057416) -- (0.66029, 0.023923) -- (0.62679, 0.033493) -- (0.58852, 0.057416) -- (0.55502, 0.047847) -- (0.58373, 0.014354) -- (0.55981, 0.0) -- (0.52153, 0.019139) -- (0.48325, 0.038278) -- (0.44498, 0.057416) -- (0.4067, 0.062201) -- (0.36842, 0.066986) -- (0.33014, 0.07177) -- (0.29187, 0.08134) -- (0.25359, 0.095694) -- (0.21531, 0.11005) -- (0.17703, 0.11005) -- (0.13876, 0.1244) -- (0.11005, 0.14833) -- (0.086124, 0.1866) -- (0.07177, 0.22488) -- (0.066986, 0.26316) -- (0.062201, 0.30144) -- (0.028708, 0.33971) -- (0.0095694, 0.37799) -- (0.0, 0.41627) -- (0.0047847, 0.45455) -- (0.019139, 0.49282) -- (0.023923, 0.50239) -- cycle;

\end{tikzpicture} }%
    \end{subfigure}%
    \hspace{.03\linewidth}%
    \begin{subfigure}[b]{0.3\linewidth}
        \resizebox{\textwidth}{!}{\begin{tikzpicture}[
    every node/.style={inner sep=0pt, outer sep=0pt, text depth=0pt, anchor=center, align=center, font=\sffamily\huge, text=black!75}]

    \foreach \x/\y in {
    0.023923 / 0.50239, 0.0382775 / 0.52153, 0.052632 / 0.54067, 0.052632 / 0.54067, 0.06698599999999999 / 0.557415, 0.08134 / 0.57416, 0.08134 / 0.57416, 0.0933 / 0.5932999999999999, 0.10526 / 0.61244, 0.10526 / 0.61244, 0.11244000000000001 / 0.63158, 0.11962 / 0.65072, 0.11962 / 0.65072, 0.133975 / 0.6698599999999999, 0.14833 / 0.689, 0.14833 / 0.689, 0.16746499999999997 / 0.696175, 0.1866 / 0.70335, 0.1866 / 0.70335, 0.20573999999999998 / 0.698565, 0.22488 / 0.69378, 0.22488 / 0.69378, 0.232055 / 0.6746399999999999, 0.23923 / 0.6555, 0.23923 / 0.6555, 0.25837 / 0.66507, 0.27751 / 0.67464, 0.27751 / 0.67464, 0.294255 / 0.688995, 0.311 / 0.70335, 0.311 / 0.70335, 0.33014 / 0.720095, 0.34928 / 0.73684, 0.34928 / 0.73684, 0.36841999999999997 / 0.74641, 0.38756 / 0.75598, 0.38756 / 0.75598, 0.4067 / 0.76555, 0.42584 / 0.77512, 0.42584 / 0.77512, 0.444975 / 0.782295, 0.46411 / 0.78947, 0.46411 / 0.78947, 0.48086 / 0.80861, 0.49761 / 0.82775, 0.49761 / 0.82775, 0.514355 / 0.8468899999999999, 0.5311 / 0.86603, 0.5311 / 0.86603, 0.5502400000000001 / 0.880385, 0.56938 / 0.89474, 0.56938 / 0.89474, 0.58134 / 0.913875, 0.5933 / 0.93301, 0.5933 / 0.93301, 0.5956950000000001 / 0.95215, 0.59809 / 0.97129, 0.59809 / 0.97129, 0.62679 / 1.0, 0.62679 / 1.0, 0.64593 / 1.0, 0.66507 / 1.0, 0.66507 / 1.0, 0.68421 / 0.9928250000000001, 0.70335 / 0.98565, 0.70335 / 0.98565, 0.720095 / 0.9760800000000001, 0.73684 / 0.96651, 0.73684 / 0.96651, 0.7559800000000001 / 0.964115, 0.77512 / 0.96172, 0.77512 / 0.96172, 0.79426 / 0.96172, 0.8134 / 0.96172, 0.8134 / 0.96172, 0.832535 / 0.95933, 0.85167 / 0.95694, 0.85167 / 0.95694, 0.86842 / 0.944975, 0.88517 / 0.93301, 0.88517 / 0.93301, 0.89474 / 0.913875, 0.90431 / 0.89474, 0.90431 / 0.89474, 0.90431 / 0.8755999999999999, 0.90431 / 0.85646, 0.90431 / 0.85646, 0.901915 / 0.8373200000000001, 0.89952 / 0.81818, 0.89952 / 0.81818, 0.8947350000000001 / 0.79904, 0.88995 / 0.7799, 0.88995 / 0.7799, 0.88995 / 0.760765, 0.88995 / 0.74163, 0.88995 / 0.74163, 0.88756 / 0.7224900000000001, 0.88517 / 0.70335, 0.88517 / 0.70335, 0.8755999999999999 / 0.68421, 0.86603 / 0.66507, 0.86603 / 0.66507, 0.85646 / 0.64593, 0.84689 / 0.62679, 0.84689 / 0.62679, 0.83493 / 0.6076550000000001, 0.82297 / 0.58852, 0.82297 / 0.58852, 0.811005 / 0.57177, 0.79904 / 0.55502, 0.79904 / 0.55502, 0.779905 / 0.54067, 0.76077 / 0.52632, 0.76077 / 0.52632, 0.75359 / 0.50957, 0.74641 / 0.49282, 0.74641 / 0.49282, 0.741625 / 0.473685, 0.73684 / 0.45455, 0.73684 / 0.45455, 0.73684 / 0.43540999999999996, 0.73684 / 0.41627, 0.73684 / 0.41627, 0.73684 / 0.39713, 0.73684 / 0.37799, 0.73684 / 0.37799, 0.7511950000000001 / 0.35885, 0.76555 / 0.33971, 0.76555 / 0.33971, 0.77512 / 0.322965, 0.78469 / 0.30622, 0.78469 / 0.30622, 0.801435 / 0.322965, 0.81818 / 0.33971, 0.81818 / 0.33971, 0.84211 / 0.32057, 0.84211 / 0.32057, 0.858855 / 0.322965, 0.8756 / 0.32536, 0.8756 / 0.32536, 0.8947400000000001 / 0.32775, 0.91388 / 0.33014, 0.91388 / 0.33014, 0.91627 / 0.313395, 0.91866 / 0.29665, 0.91866 / 0.29665, 0.8995200000000001 / 0.29186500000000004, 0.88038 / 0.28708, 0.88038 / 0.28708, 0.8756 / 0.26316, 0.8756 / 0.26316, 0.8947400000000001 / 0.26316, 0.91388 / 0.26316, 0.91388 / 0.26316, 0.9378 / 0.23923, 0.9378 / 0.23923, 0.921055 / 0.23445, 0.90431 / 0.22967, 0.90431 / 0.22967, 0.88517 / 0.224885, 0.86603 / 0.2201, 0.86603 / 0.2201, 0.8468899999999999 / 0.205745, 0.82775 / 0.19139, 0.82775 / 0.19139, 0.80861 / 0.188995, 0.78947 / 0.1866, 0.78947 / 0.1866, 0.770335 / 0.1866, 0.7512 / 0.1866, 0.7512 / 0.1866, 0.73445 / 0.198565, 0.7177 / 0.21053, 0.7177 / 0.21053, 0.70574 / 0.227275, 0.69378 / 0.24402, 0.69378 / 0.24402, 0.66507 / 0.27273, 0.66507 / 0.27273, 0.6531100000000001 / 0.291865, 0.64115 / 0.311, 0.64115 / 0.311, 0.61722 / 0.31579, 0.61722 / 0.31579, 0.600475 / 0.299045, 0.58373 / 0.2823, 0.58373 / 0.2823, 0.5645899999999999 / 0.270335, 0.54545 / 0.25837, 0.54545 / 0.25837, 0.526315 / 0.251195, 0.50718 / 0.24402, 0.50718 / 0.24402, 0.50239 / 0.21531, 0.50239 / 0.21531, 0.52632 / 0.19139, 0.52632 / 0.19139, 0.545455 / 0.188995, 0.56459 / 0.1866, 0.56459 / 0.1866, 0.569375 / 0.16985499999999998, 0.57416 / 0.15311, 0.57416 / 0.15311, 0.5932999999999999 / 0.14354, 0.61244 / 0.13397, 0.61244 / 0.13397, 0.6244000000000001 / 0.117225, 0.63636 / 0.10048, 0.63636 / 0.10048, 0.61244 / 0.08134, 0.61244 / 0.08134, 0.63158 / 0.069378, 0.65072 / 0.057416, 0.65072 / 0.057416, 0.655505 / 0.0406695, 0.66029 / 0.023923, 0.66029 / 0.023923, 0.64354 / 0.028708, 0.62679 / 0.033493, 0.62679 / 0.033493, 0.6076550000000001 / 0.0454545, 0.58852 / 0.057416, 0.58852 / 0.057416, 0.57177 / 0.0526315, 0.55502 / 0.047847, 0.55502 / 0.047847, 0.569375 / 0.0311005, 0.58373 / 0.014354, 0.58373 / 0.014354, 0.55981 / 0.0, 0.55981 / 0.0, 0.54067 / 0.0095695, 0.52153 / 0.019139, 0.52153 / 0.019139, 0.50239 / 0.028708499999999998, 0.48325 / 0.038278, 0.48325 / 0.038278, 0.464115 / 0.047847, 0.44498 / 0.057416, 0.44498 / 0.057416, 0.42584 / 0.0598085, 0.4067 / 0.062201, 0.4067 / 0.062201, 0.38756 / 0.0645935, 0.36842 / 0.066986, 0.36842 / 0.066986, 0.34928000000000003 / 0.069378, 0.33014 / 0.07177, 0.33014 / 0.07177, 0.311005 / 0.076555, 0.29187 / 0.08134, 0.29187 / 0.08134, 0.27273000000000003 / 0.088517, 0.25359 / 0.095694, 0.25359 / 0.095694, 0.23445 / 0.10287199999999999, 0.21531 / 0.11005, 0.21531 / 0.11005, 0.19617 / 0.11005, 0.17703 / 0.11005, 0.17703 / 0.11005, 0.157895 / 0.117225, 0.13876 / 0.1244, 0.13876 / 0.1244, 0.11005 / 0.14833, 0.11005 / 0.14833, 0.09808700000000001 / 0.16746499999999997, 0.086124 / 0.1866, 0.086124 / 0.1866, 0.078947 / 0.20573999999999998, 0.07177 / 0.22488, 0.07177 / 0.22488, 0.069378 / 0.24402000000000001, 0.066986 / 0.26316, 0.066986 / 0.26316, 0.0645935 / 0.2823, 0.062201 / 0.30144, 0.062201 / 0.30144, 0.0454545 / 0.320575, 0.028708 / 0.33971, 0.028708 / 0.33971, 0.0191387 / 0.35885, 0.0095694 / 0.37799, 0.0095694 / 0.37799, 0.0047847 / 0.39713, 0.0 / 0.41627, 0.0 / 0.41627, 0.00239235 / 0.43540999999999996, 0.0047847 / 0.45455, 0.0047847 / 0.45455, 0.01196185 / 0.473685, 0.019139 / 0.49282, 0.019139 / 0.49282, 0.023923 / 0.50239
    }
    {
        \draw [vll-dark, fill=vll-orange, line width=.1pt] (\x,\y) circle (.2pt);
    }

\end{tikzpicture} }%
    \end{subfigure}%
    \hspace{.03\linewidth}%
    \begin{subfigure}[b]{0.3\linewidth}
        \resizebox{\textwidth}{!}{\input{figures/fourier-curves} }%
    \end{subfigure}%
    \caption{\emph{Left:} A 2d polygon obtained from the segmentation of a natural image.
    \emph{Middle:} The vertices $\mathbf{p}_l$ of the polygon forming the basis for computing the Fourier coefficients in \cref{eq:fourier-coefficients}.
    \emph{Right:} Tracing the curve $\mathbf{g}(t)$ according to \cref{eq:fourier-curve}, for different numbers $M$ of Fourier terms $\mathbf{a}_m$.}
    \label{fig:fourier-shape}
\end{figure}

\begin{figure*}[t]
    \centering
    \begin{subfigure}{0.49\linewidth}
        \resizebox{\textwidth}{!}{\includegraphics[]{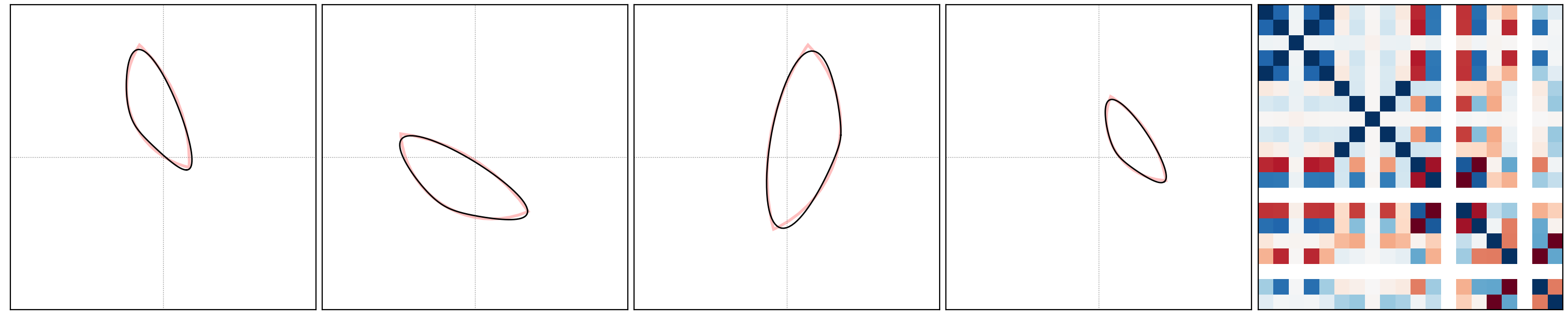} }%
    \end{subfigure}%
    \hspace{.01\linewidth}%
    \begin{subfigure}{0.49\linewidth}
        \resizebox{\textwidth}{!}{\includegraphics[]{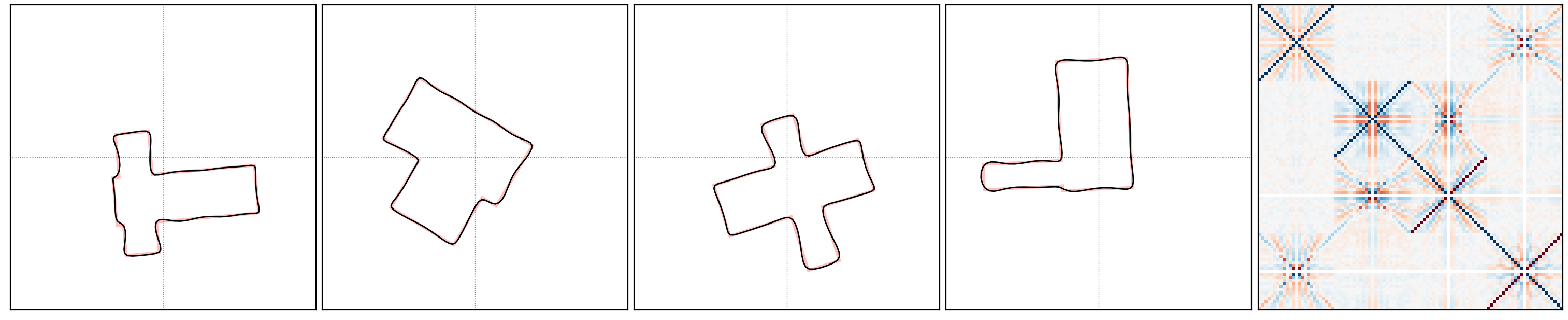} }%
    \end{subfigure}%
    \caption{\emph{Left:} Samples from the \emph{Lens} shapes data set, and true correlation matrix of Fourier coefficients for a large batch of \emph{Lens} shapes.
    \emph{Right:} The same for \emph{Cross} shapes.
    Red lines behind the Fourier curves show the original geometry they approximate.}
    \label{fig:data-sets}
\end{figure*}

\begin{figure*}[t]
    \centering
    \begin{subfigure}{0.49\linewidth}
        \resizebox{\textwidth}{!}{\includegraphics[]{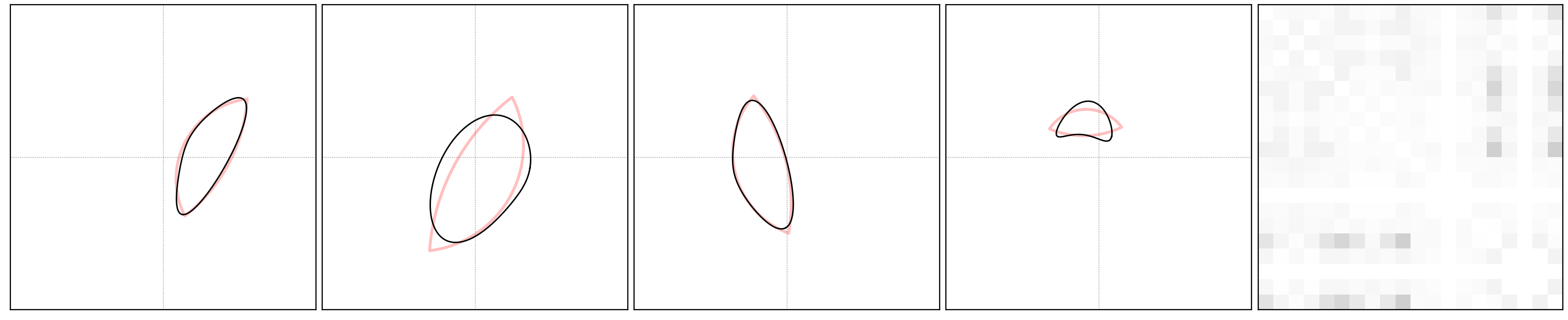} }%
    \end{subfigure}%
    \hspace{.01\linewidth}%
    \begin{subfigure}{0.49\linewidth}
        \resizebox{\textwidth}{!}{\includegraphics[]{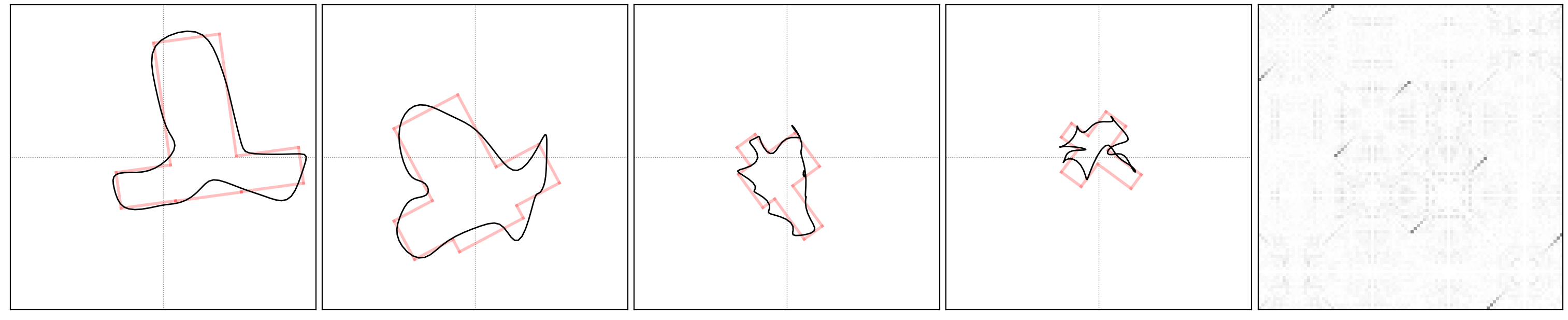} }%
    \end{subfigure}\\
    \begin{subfigure}{0.49\linewidth}
        \resizebox{\textwidth}{!}{\includegraphics[]{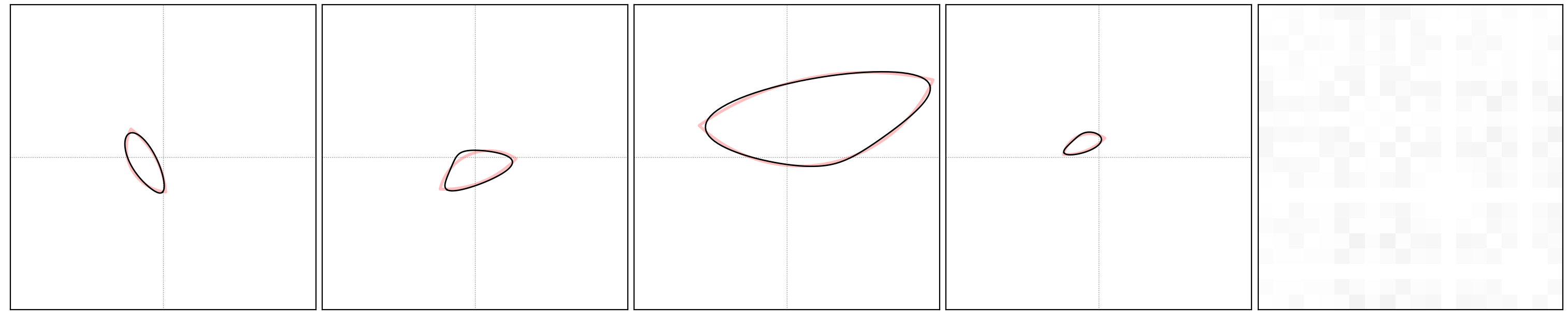} }%
    \end{subfigure}%
    \hspace{.01\linewidth}%
    \begin{subfigure}{0.49\linewidth}
        \resizebox{\textwidth}{!}{\includegraphics[]{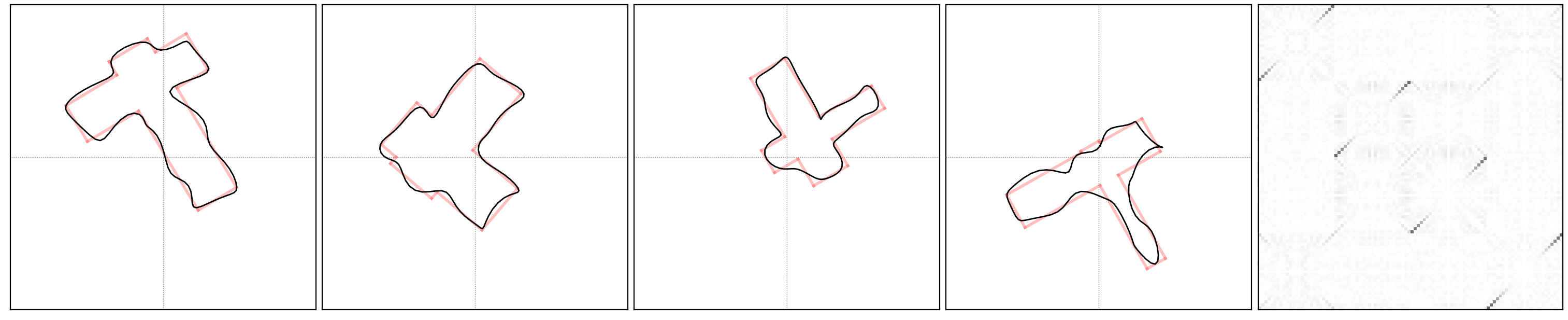} }%
    \end{subfigure}%
    \caption{Samples from \textsc{Real-NVP} \emph{(top)} and \textsc{recursive} \emph{(bottom)} coupling block networks,
    trained on the \emph{Lens} shapes \emph{(left)} and the \emph{Cross} shapes \emph{(right)} from \cref{fig:data-sets}.
    The closest fitting shapes from the true distributions are depicted in red for reference.
    Here and in subsequent figures, the fifth panel shows the absolute differences between true and sampling correlation matrices.}
    \label{fig:unconditional-samples}
\end{figure*}

We perform experiments on two  specific data sets, first using curves of order $M=2$, i.e.~$|\x| = 20$, to represent a distribution of simple shapes that arise from the intersection of two randomly placed circles with a fixed ratio of radii and a fixed distance.
The resulting \emph{Lens} shapes can be seen in \cref{fig:data-sets} \emph{(left)}, together with the highly structured correlation matrix of Fourier parameters $\x_i$ that yield such shapes.

The second data set uses $M=12$, i.e.~$|\x| = 100$, to represent \emph{Cross} shapes which are generated by crossing two bars of random length, width and lateral shift at a right angle, oriented randomly, but positioned close to the origin.
This results in a variation of \textsf{X}s, \textsf{L}s and \textsf{T}s, some of which are shown in \cref{fig:data-sets} \emph{(right)} together with the even more complicated ($100 \times 100$) parameter correlation matrix.


\subsubsection{Density estimation.}

For density estimation, a single-block and a two-block network are trained  on the \emph{Lens}-shapes data, once with standard coupling blocks and once with the new recursive design.
All networks have the same total parameter budget -- details in the appendix.


Samples from the two-block models and absolute differences to the true parameter correlation matrices are shown in \cref{fig:unconditional-samples} \emph{(left)}.
Qualitatively, samples from the recursive model are visually more faithful 
and have smaller errors in the correlation matrices.
Quantitatively, we compare over three training runs per model using the following metrics:
\begin{itemize}
    \item \textbf{Maximum mean discrepancy} \citep[\textsc{MMD},][]{gretton2012kernel} measures the dissimilarity of two distributions using only samples from both.
    Following \cite{ardizzone2018analyzing}, we use MMD with an inverse multi-quadratic kernel and average the results over $100$ batches from the data prior and from each trained model.
    Lower is better.
    \item Average \textbf{log-likelihood} (\textsc{LL}) of the test data under the model, i.e. $-\tfrac{1}{2} T(\x)^2 + \text{log}|\mathbf{J}_T(\x)| - \text{log}( 2\pi \tfrac{N}{2})$ where $N$ is the data dimensionality.
    Higher is better.
    \item Average \textbf{intersection-over-union} (\textsc{IoU}) between generated shapes and the best fitting shape that follows the construction rules of the data set.
    See appendix for details on the fitting procedure.
    Higher is better.
    \item Average \textbf{Hausdorff distance} (\textsc{H-dist}) between the contours of the generated shapes and those of the best fitting shapes, as above.
    Lower is better.
\end{itemize}

\Cref{tab:density-estimation-small} shows how recursive coupling blocks outperform the conventional design.
The difference is especially striking for the single-block network, as a non-recursive coupling block leaves half the variables untouched and is thus inherently unable to model the data distribution properly.

We also trained standard and recursive networks with 4 and 8 coupling blocks on the larger \emph{Cross}-shapes data set.
Representative samples from the 4-block models are shown in \cref{fig:unconditional-samples} \emph{(right)} together with the best fitting, actual \emph{Cross} shape.
Here, it is even more clearly visible that the recursive model produces samples with better geometry, i.e~right angles, straight lines and symmetries in the expected locations.

A quantitative comparison in terms of \textsc{LL}, \textsc{IoU} and \textsc{H-dist} is presented in \cref{tab:unconditional-plus}, with recursive coupling consistently outperforming standard \textsc{Real-NVP} blocks.

\begin{figure*}[t]
    \centering
    \begin{subfigure}{0.49\linewidth}
        \resizebox{\textwidth}{!}{\includegraphics[]{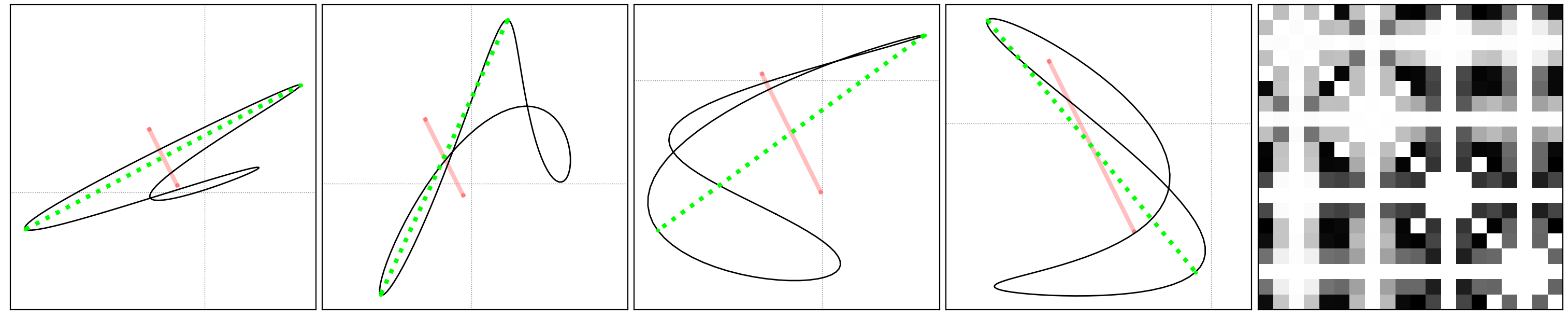} }%
    \end{subfigure}%
    \hspace{.01\linewidth}%
    \begin{subfigure}{0.49\linewidth}
        \resizebox{\textwidth}{!}{\includegraphics[]{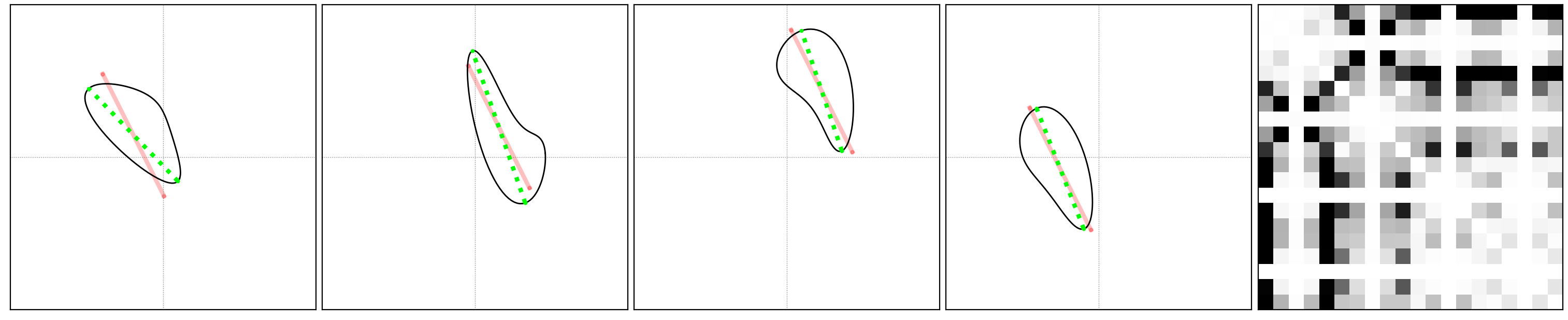} }%
    \end{subfigure}\\
    \begin{subfigure}{0.49\linewidth}
        \resizebox{\textwidth}{!}{\includegraphics[]{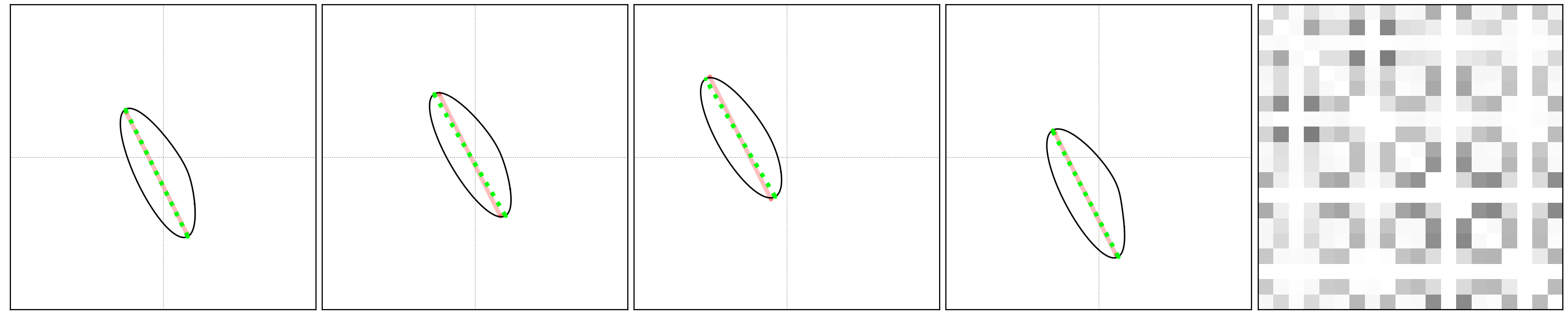} }%
    \end{subfigure}%
    \hspace{.01\linewidth}%
    \begin{subfigure}{0.49\linewidth}
        \resizebox{\textwidth}{!}{\includegraphics[]{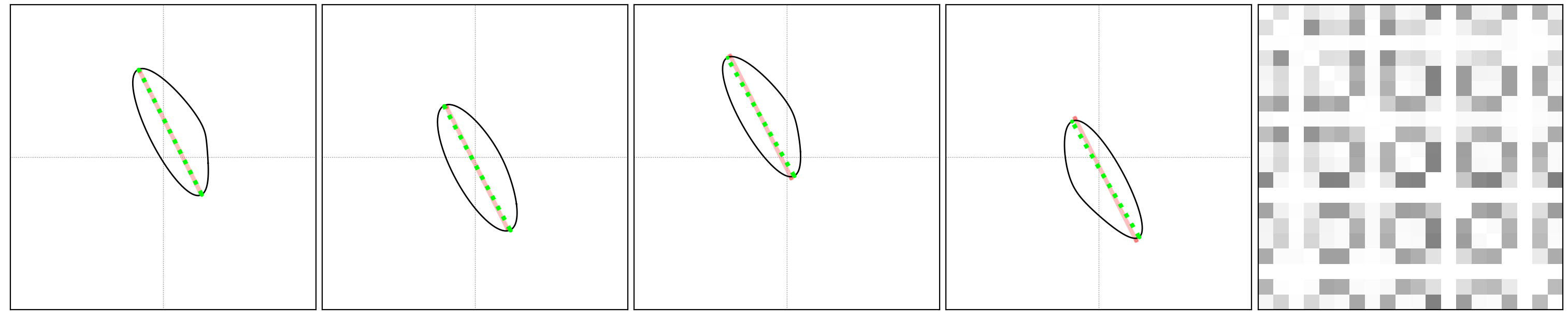} }%
    \end{subfigure}%
    \caption{Samples from a conditional coupling net \emph{(\textsc{cINN}, left)} and from \textsc{HINT} \emph{(right)}, trained \emph{Lens} shapes.
    Green dotted lines mark the largest diameter of each shape; red lines show how it should look according to the data $\y$.
    Both models do well with 4 blocks \emph{(bottom)}, but only HINT generates reasonable samples with a single coupling block \emph{(top)}.
    Last panels as in \cref{fig:unconditional-samples}.}
    \label{fig:conditional-samples-gaussian}
\end{figure*}

\begin{figure*}[t]
    \centering
    \begin{subfigure}{0.49\linewidth}
        \resizebox{\textwidth}{!}{\includegraphics[]{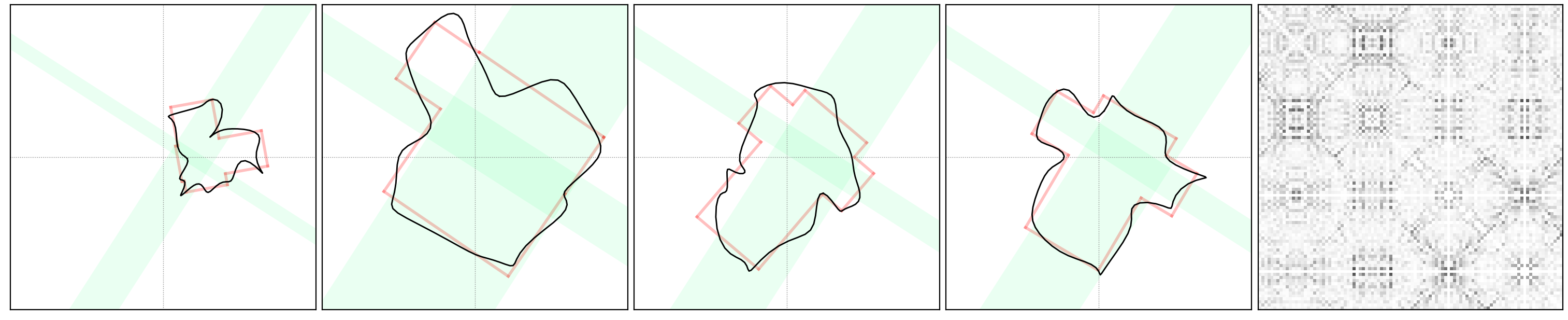} }%
    \end{subfigure}%
    \hspace{.01\linewidth}%
    \begin{subfigure}{0.49\linewidth}
        \resizebox{\textwidth}{!}{\includegraphics[]{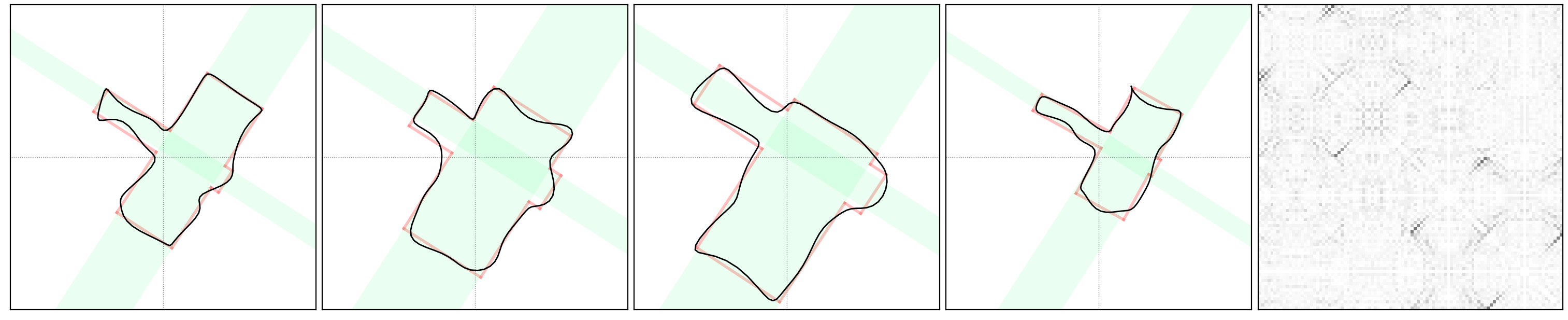} }%
    \end{subfigure}\\
    \begin{subfigure}{0.49\linewidth}
        \resizebox{\textwidth}{!}{\includegraphics[]{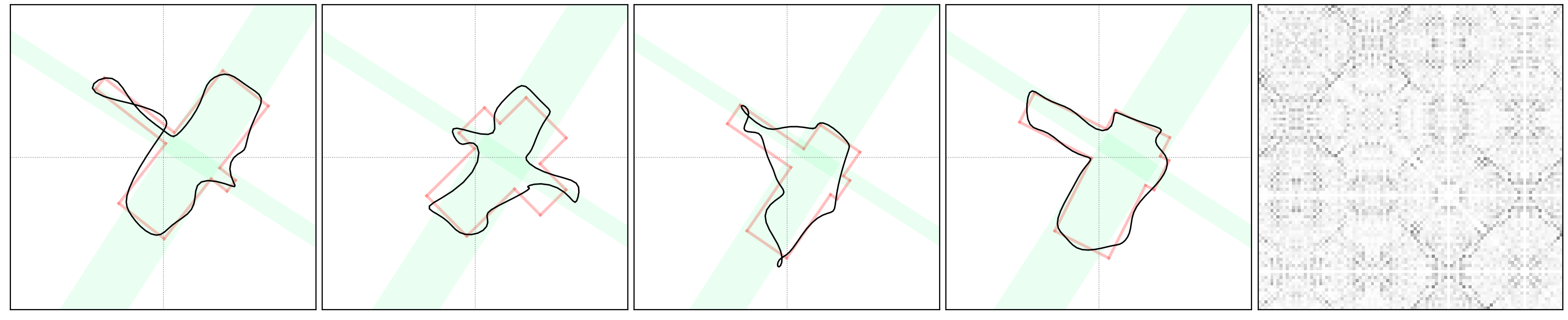} }%
    \end{subfigure}%
    \hspace{.01\linewidth}%
    \begin{subfigure}{0.49\linewidth}
        \resizebox{\textwidth}{!}{\includegraphics[]{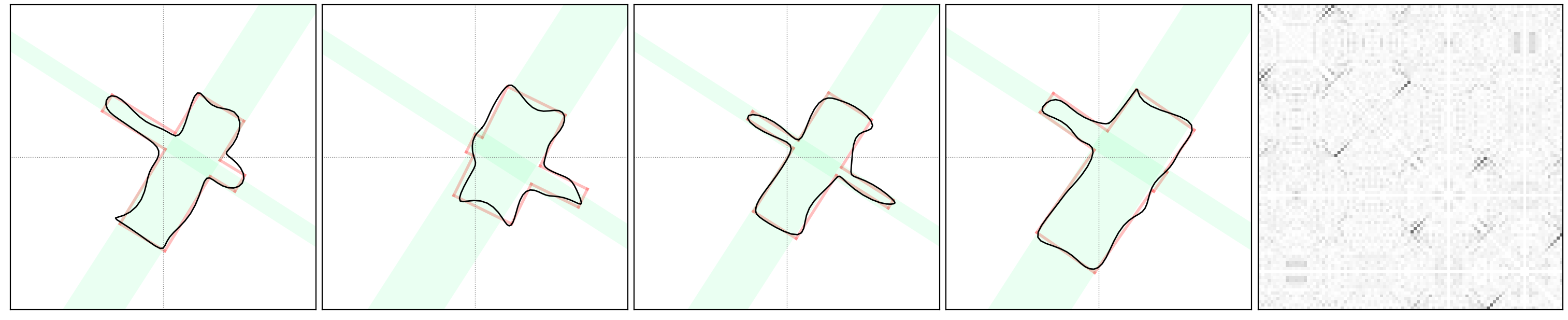} }%
    \end{subfigure}%
    \caption{Samples from a conditional coupling net \emph{(\textsc{cINN}, left)} and from \textsc{HINT} \emph{(right)} with 4 \emph{(top)} or 8 \emph{(bottom)} blocks, trained on \emph{Cross} shapes.
    The expected center, angle and thickness ratio of the \emph{Cross} according to the data $\y$ are shown in green, the best fitting \emph{Cross} shape in red.
    Visually, HINT reproduces shapes from the data set much better.
    Last panels as in \cref{fig:unconditional-samples}.}
    \label{fig:conditional-samples-plus}
\end{figure*}

\begin{table}[t]
    \begin{center}\begin{small}\begin{sc}
    \begin{tabular}{l l@{\quad}
        S[detect-weight,mode=text,table-format=1.3(4)]@{\quad}
        S[detect-weight,mode=text,table-format=1.3(4)]}
    \toprule
    Blocks & &  {Real-NVP} & {Recursive} \\
    \midrule
    & {MMD $\downarrow$} & 0.370 \pm 0.000 & \B 0.140 \pm 0.002 \\
    1 & {LL $\uparrow$} & 2.021 \pm 0.001 & \B 2.861 \pm 0.006 \\
    & {IoU $\uparrow$} & 0.449 \pm 0.010 & \B 0.688 \pm 0.014 \\
    & {H-dist $\downarrow$} & 0.519 \pm 0.005 & \B 0.109 \pm 0.005 \\
    \midrule
    & {MMD $\downarrow$} & 0.012 \pm 0.001 & \B 0.009 \pm 0.004 \\
    2 & {LL $\uparrow$} & 3.141 \pm 0.036 & \B 3.219 \pm 0.005 \\
    & {IoU $\uparrow$} & 0.789 \pm 0.024 & \B 0.819 \pm 0.006 \\
    & {H-dist $\downarrow$} & 0.063 \pm 0.007 & \B 0.057 \pm 0.000 \\
    \bottomrule
    \end{tabular}
    \end{sc}\end{small}\end{center}
    \caption{Comparing \textsc{Real-NVP} and \textsc{recursive} coupling for sampling and density estimation for the \emph{Lens}-shapes data
    (plotting mean $\pm$ standard deviation over 3 training runs).}
    \label{tab:density-estimation-small}
\end{table}

\subsection{Bayesian Inference on Fourier Shapes}


To set-up Bayesian inference tasks, we formulate forward mappings $\x \rightarrow \y$ from $\x$ to observable features $\y$. 
Since these features are incomplete shape descriptors, the inverse $\y \rightarrow \x$ is ambiguous, and $p(\x \,|\, \y)$ is learned with HINT.

Given a \emph{Lens} shape, our forward mapping locates its two tips and returns their horizontal and vertical distances $d_h$ and $d_v$. The tips' absolute positions and the side of the lens' ``bulge'' remain undetermined by these features.

The forward mapping for \emph{Cross} shapes returns four geometrical features.
These are the 2d coordinates of the center, i.e~where the bars cross,
plus the angle of and thickness ratio between the two bars.
What remains free, are the absolute thickness, as well as the length and lateral shift of the bars.

Finally, noise $\sigma \sim \mathcal{N}(\mathbf{0}, \tfrac{1}{20} \mathbf{I})$ is added to the output of each forward mapping to obtain observed data vectors $\y$.

We trained a conditional flow model (cINN) and HINT with 1, 2, 4 and 8 blocks for Bayesian inference on the \emph{Lens} shapes.
A quantitative comparison, in terms of \textsc{MMD}, 
\textsc{IoU} and \textsc{H-dist}, is given in \cref{tab:bayesian-inference-small}.
Here, however, \textsc{MMD} does not compare to samples from the prior $p_X(\x)$, but to samples from an estimate of the true posterior $p(\x \,|\, \y)$, estimated 
via \textit{Approximate Bayesian Computation} \citep[ABC,][]{csillery2010abc}, as in \cite{ardizzone2018analyzing}, see appendix.

\begin{table}[t]
    \begin{center}\begin{small}\begin{sc}
    \begin{tabular}{l@{\qquad}
        S[detect-weight,mode=text,table-format=1.3]@{\quad}
        S[detect-weight,mode=text,table-format=1.3]@{\qquad}
        S[detect-weight,mode=text,table-format=1.3]@{\quad}
        S[detect-weight,mode=text,table-format=1.3]}
    \toprule
    Blocks & {$4 \times$ rnvp} & {$4 \times$ rec} & {$8 \times$ rnvp} & {$8 \times$ rec} \\
    \midrule
    LL $\uparrow$       & 3.419 & \B 3.627 & 3.329 & \B 3.637 \\
    IoU $\uparrow$      & 0.594 & \B 0.823 & 0.588 & \B 0.823 \\
    H-dist $\downarrow$ & 0.134 & \B 0.077 & 0.138 & \B 0.073 \\
    \bottomrule
    \end{tabular}
    \end{sc}\end{small}\end{center}
    \caption{Comparison of flow models for \emph{Cross} shapes with normal (\textsc{rnvp}) and recursive coupling (\textsc{rec}), respectively.}
    \label{tab:unconditional-plus}
\end{table}

\begin{table}[t!]
    \begin{center}\begin{small}\begin{sc}
    \begin{tabular}{l@{\quad} l@{\quad}
        S[detect-weight,mode=text,table-format=2.3(4)]@{\quad}
        S[detect-weight,mode=text,table-format=2.3(4)]}
    \toprule
    Blocks & & {cINN} & {HINT} \\
    \midrule
    & {MMD $\downarrow$}    & 0.746 \pm 0.000 & \B 0.030 \pm 0.001 \\
    1 & {IoU $\uparrow$}      & 0.456 \pm 0.001 & \B 0.849 \pm 0.003 \\
    & {H-dist $\downarrow$} & 0.521 \pm 0.006 & \B 0.051 \pm 0.001 \\
    \midrule
    & {MMD $\downarrow$}    & 0.048 \pm 0.012 & \B 0.016 \pm 0.001 \\
    2 & {IoU $\uparrow$}      & 0.839 \pm 0.011 & \B 0.869 \pm 0.005 \\
    & {H-dist $\downarrow$} & 0.054 \pm 0.002 & \B 0.044 \pm 0.002 \\
    \midrule
    & {MMD $\downarrow$}    & 0.020 \pm 0.003 & \B 0.010 \pm 0.001 \\
    4 & {IoU $\uparrow$}      & 0.865 \pm 0.006 & \B 0.875 \pm 0.007 \\
    & {H-dist $\downarrow$} & 0.046 \pm 0.002 & \B 0.043 \pm 0.002 \\
    \midrule
    & {MMD $\downarrow$}    & \B 0.007 \pm 0.002 & 0.011 \pm 0.003 \\
    8 & {IoU $\uparrow$}      & 0.864 \pm 0.003 & \B 0.876 \pm 0.003 \\
    & {H-dist $\downarrow$} & 0.046 \pm 0.001 & \B 0.043 \pm 0.001 \\
    \bottomrule
    \end{tabular}
    \end{sc}\end{small}\end{center}
    \caption{Conditional (\textsc{cINN}) vs.~hierarchical (\textsc{HINT}) coupling for Bayesian inference on \emph{Lens} shapes
    (mean $\pm$ standard deviation over 3 training runs).
    See text for details.}
    \label{tab:bayesian-inference-small}
\end{table}

In \cref{tab:bayesian-inference-small} we see that
HINT consistently produces better shapes (measured by \textsc{IoU} and \textsc{H-dist}), especially in the case of a single block, and it exhibits better conditioning (as evidenced by \textsc{MMD}) 
for all but the 8-block model, most likely due to the limited parameter budget, which leaves some of the sub-networks  underparameterized. A similar effect can be observed in the \textsc{LL} (not shown in \cref{tab:bayesian-inference-small}), when only measuring the log-likelihood of the $\x$-lane in \textsc{HINT} and simply excluding contributions from the $\y$-lane.

Qualitative results in figures \ref{fig:conditional-samples-gaussian} and \ref{fig:conditional-samples-plus}, for the \textit{Lens} and for the \textit{Cross} shapes, 
respectively, confirm the superior performance of \textsc{HINT} also visually, in particular 
%
producing significantly better angles and symmetries for the \textit{Cross} shapes. 
%
This is quantitatively supported in the metrics in \cref{tab:conditional-plus}.
%
\begin{table}[t]
    \begin{center}\begin{small}\begin{sc}
    \begin{tabular}{l@{\quad}
        S[detect-weight,mode=text,table-format=1.3]@{\quad}
        S[detect-weight,mode=text,table-format=1.3]@{\qquad}
        S[detect-weight,mode=text,table-format=1.3]@{\quad}
        S[detect-weight,mode=text,table-format=1.3]}
    \toprule
    Blocks & {$4 \times\!$ cINN} & {$4 \times\!$ HINT} & {$8 \times\!$ cINN} & {$8 \times\!$ HINT} \\
    \midrule
    LL $\uparrow$       & 3.625 & \B 3.724 & 3.609 & \B 3.766 \\
    IoU $\uparrow$      & 0.654 & \B 0.843 & 0.590 & \B 0.859 \\
    H-dist $\downarrow$ & 0.116 & \B 0.073 & 0.119 & \B 0.066 \\
    \bottomrule
    \end{tabular}
    \end{sc}\end{small}\end{center}
    \caption{Fidelity of \emph{Cross} shapes generated by conditional (\textsc{cINN}) and hierarchical (\textsc{HINT}) flows, respectively.}
    \label{tab:conditional-plus}
\end{table}

\subsection{Recursion Depth vs.~Number of Blocks}
\label{sec:depth-tradeoff}

Seeing that recursive coupling blocks outperform standard ones, we also tested if using more standard coupling blocks closes the gap.
We looked at several combinations of number of blocks, recursion depth and parameter budget on the \emph{Cross}-shapes data.
In summary, the first two recursion levels improve performance more than additional coupling blocks.
Beyond that, we see diminishing returns, as the limited parameter budget gets distributed over too many subnetworks.
Full results are in the appendix.



\section{Conclusion}
\label{sec:conclusion}

We presented recursive coupling blocks and HINT, a new invertible architecture for 
normalizing flow, improving on the traditional coupling block in terms of expressive power by densifying  the triangular Jacobian, while keeping the advantages of an accessible latent space. 
This keeps the efficient sampling and density estimation of RealNVP, which is often compromised by other approaches to denser Jacobians, e.g. auto-regressive flows.
To evaluate the model, we introduced a versatile family of data sets based on Fourier decompositions of simple 2D shapes that can be visualized easily, independent of the chosen dimension.
In terms of future improvements, we expect that our formulation can be made more computationally efficient
through the use of e.g.~masking operations, enabling more advanced parallelization.

\begin{figure*}
    \centering
    {\LARGE\bf -- APPENDIX -- \par}
    \vspace{2em}
\end{figure*}
\FloatBarrier

\begin{figure}
    \centering
    \resizebox{0.8\linewidth}{!}{\begin{tikzpicture}[
    on layer/.code={
        \pgfonlayer{#1}\begingroup
        \aftergroup\endpgfonlayer
        \aftergroup\endgroup
    },%
    draw on back/.style={
        preaction={
            draw,
            on layer=back,
            line width=2pt
        },
        transform shape
    }]

    \begin{scope}[shift={(0, 0)}, rotate=-10]
        \drawfill (-1, .8) rectangle (1, .2);
        \drawfill (.6, -.8) rectangle (1, .8);
    \end{scope}

    \begin{scope}[shift={(2.5, 0)}, rotate=-10]
        \drawfill (-1, .8) rectangle (1, .2);
        \drawfill (-.2, -.8) rectangle (.2, .8);
    \end{scope}

    \begin{scope}[shift={(5, 0)}, rotate=-10]
        \drawfill (-1, .8) rectangle (1, .2);
        \drawfill (-1, -.8) rectangle (-.6, .8);
    \end{scope}

    \begin{scope}[shift={(0, 2.5)}, rotate=-10]
        \drawfill (-1, -.3) rectangle (1, .3);
        \drawfill (.6, -.8) rectangle (1, .8);
    \end{scope}

    \begin{scope}[shift={(2.5, 2.5)}, rotate=-10]
        \drawfill (-1, -.3) rectangle (1, .3);
        \drawfill (-.2, -.8) rectangle (.2, .8);
    \end{scope}

    \begin{scope}[shift={(5, 2.5)}, rotate=-10]
        \drawfill (-1, -.3) rectangle (1, .3);
        \drawfill (-1, -.8) rectangle (-.6, .8);
    \end{scope}

    \begin{scope}[shift={(0, 5)}, rotate=-10]
        \drawfill (-1, -.8) rectangle (1, -.2);
        \drawfill (.6, -.8) rectangle (1, .8);
    \end{scope}

    \begin{scope}[shift={(2.5, 5)}, rotate=-10]
        \drawfill (-1, -.8) rectangle (1, -.2);
        \drawfill (-.2, -.8) rectangle (.2, .8);
    \end{scope}

    \begin{scope}[shift={(5, 5)}, rotate=-10]
        \drawfill (-1, -.8) rectangle (1, -.2);
        \drawfill (-1, -.8) rectangle (-.6, .8);
    \end{scope}

\end{tikzpicture} }
    \caption{Configurations for initializing the shape fitting algorithm on \emph{Cross} shapes, with a given angle.}
    \label{fig:fit-plus-shapes}
\end{figure}

\begin{figure}
    \captionsetup[subfigure]{labelformat=empty,singlelinecheck=off,justification=raggedright,skip=0pt}
    \centering
    \begin{subfigure}{.98\linewidth}
        \caption{\textsc{Real-NVP}, 8 blocks:}
        \resizebox{\textwidth}{!}{\includegraphics[]{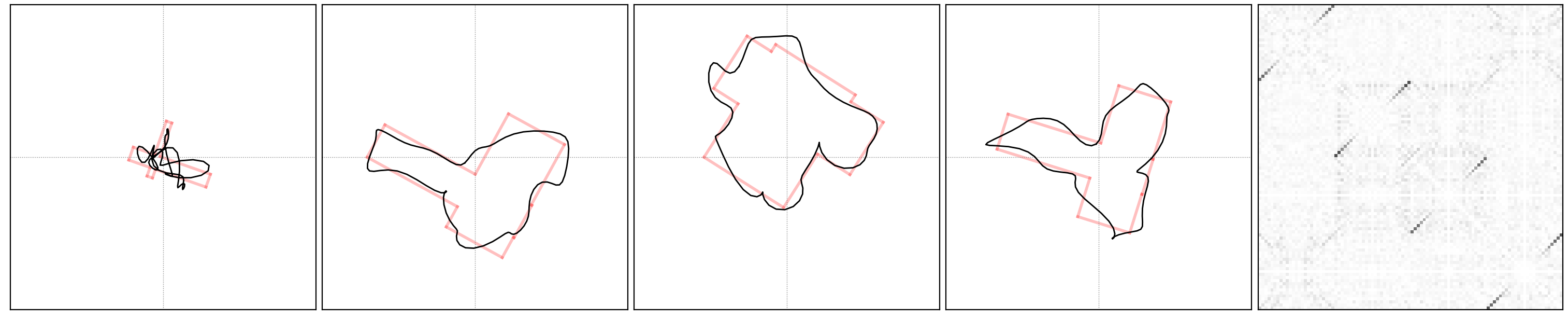} }%
    \end{subfigure}\\%
    \begin{subfigure}{.98\linewidth}
        \caption{\textsc{Real-NVP}, 16 blocks:}
        \resizebox{\textwidth}{!}{\includegraphics[]{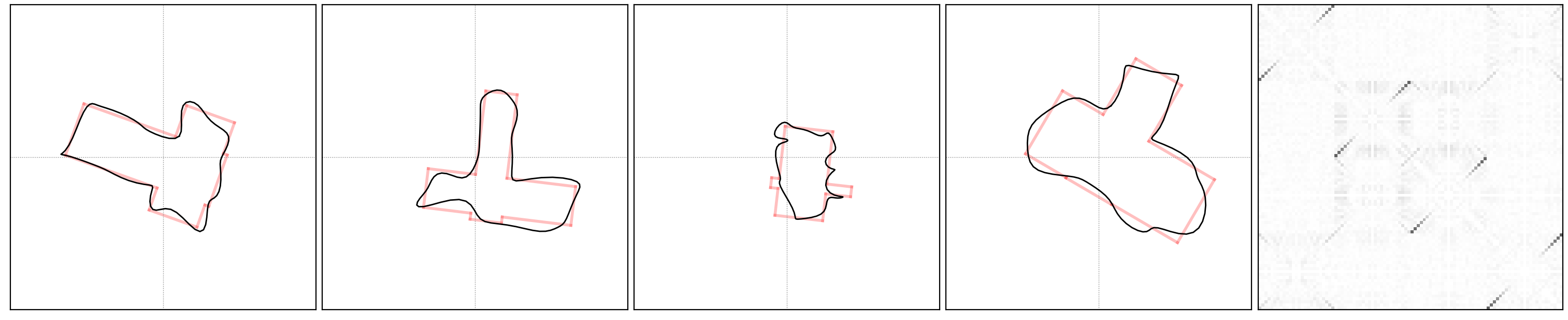} }%
    \end{subfigure}\\%
    \begin{subfigure}{.98\linewidth}
        \caption{\textsc{Real-NVP}, 32 blocks:}
        \resizebox{\textwidth}{!}{\includegraphics[]{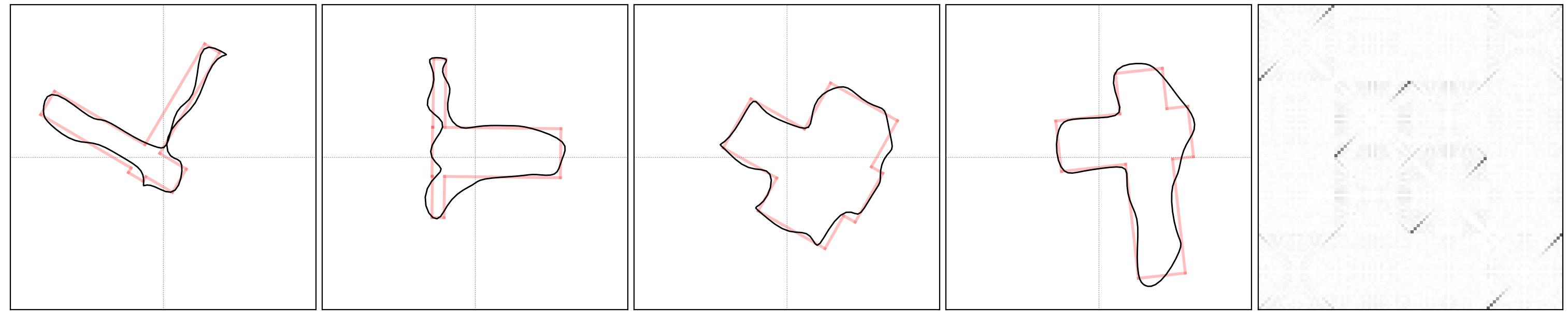} }%
    \end{subfigure}\\%
    \begin{subfigure}{.98\linewidth}
        \caption{\textsc{Recursive coupling}, 4 blocks:}
        \resizebox{\textwidth}{!}{\includegraphics[]{figures/samples/plus-shape_unconditional_hint-4-full_example.png} }%
    \end{subfigure}%
    \caption{Samples from deeper \textsc{Real-NVP} models vs.~ours, visualized in the same style as in the main paper.}
    \label{fig:deep-samples-plus}
    \vspace{4em}

    \resizebox{\linewidth}{!}{ \includegraphics{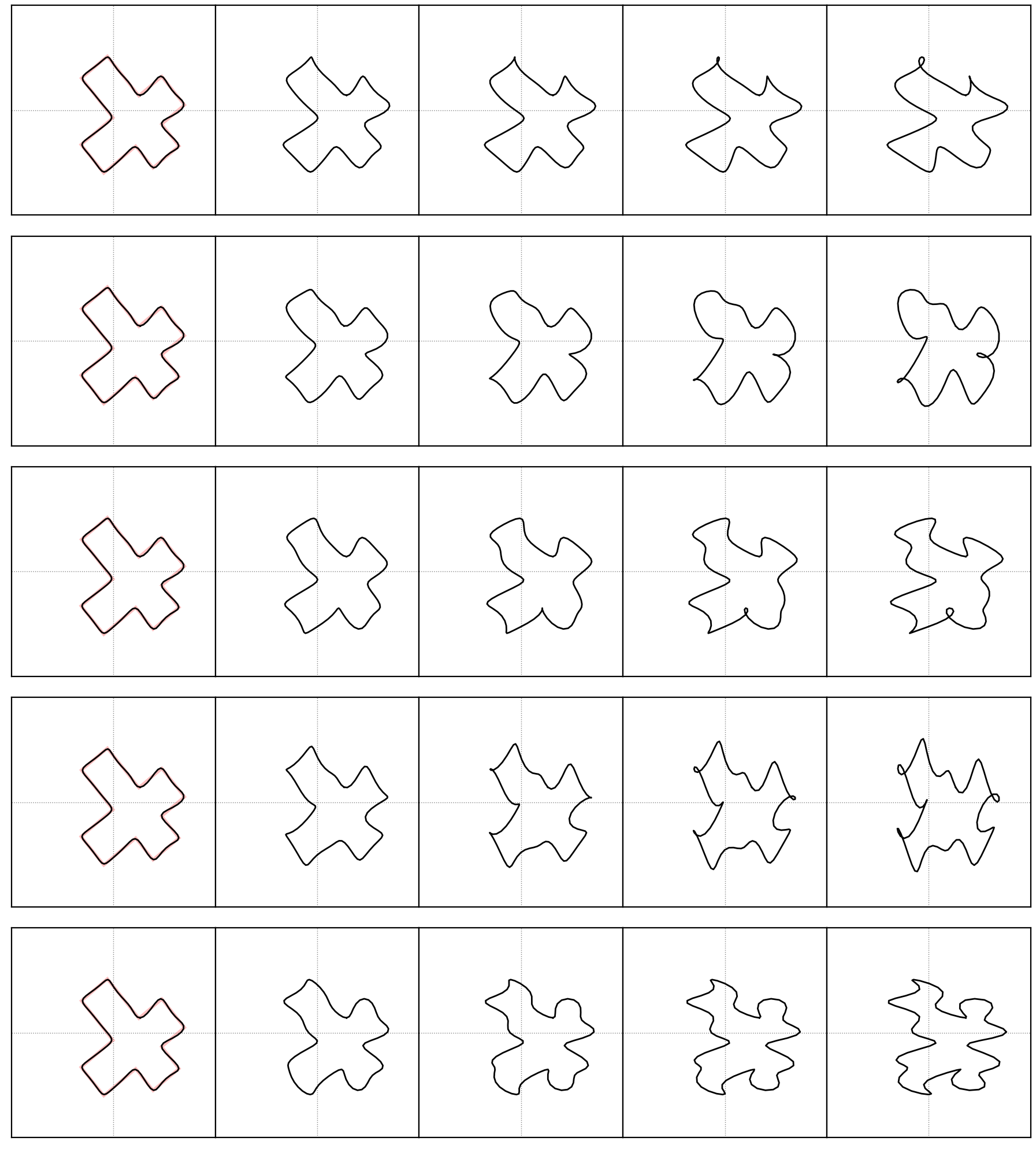} }
    \caption{Sensitivity of Fourier shapes to changes in a single parameter.}
    \label{fig:parameter-sensitivity}
\end{figure}

\section{Details on \emph{Lens} Shapes Data Set}
\label{sec:appendix-lens}

The shapes in the \emph{Lens} data set arise from the intersection of two circles.
The first has a uniformly random radius $r$ within $[1,2]$, the second has exactly double that radius.
They are positioned at a uniformly random angle, with their centers set apart by a distance of $2.4r$.
Their intersection is centered at the origin and offset by a random distance from $\mathcal{N}(0,\tfrac{1}{2})$ in either dimension.

\section{Details on \emph{Cross} Shapes Data Set}
\label{sec:appendix-cross}

The shapes for the larger \emph{Cross} data set are generated by taking the union of two oblong rectangles crossing each other at a right angle.
For both rectangles, the longer side length is drawn uniformly from $[\,3,5\,]$ and the other from $[\,\tfrac{1}{2},2\,]$.
We shift both rectangles along their longer side by a uniformly random distance drawn from $[\,-\tfrac{3}{2},\tfrac{3}{2}\,]$.

Then we form the union and insert equally spaced points along the resulting polygon's sides such that no line segment is longer than $\tfrac{1}{5}$.
This is necessary to obtain dense point sequences which are approximated more faithfully by Fourier curves.

Finally, we center the shape at the origin, rotate it by a random angle and again shift it within the plane by a distance drawn from $\mathcal{N}(0,\tfrac{1}{2})$.


\section{Training Hyper-Parameters}
\label{sec:appendix-hyperparams}

For all experiments, we used the Adam optimizer with betas $(0.9, 0.95)$ for $50$ epochs with L2 weight regularization strength $1.86 \cdot 10^{-5}$.
Unless otherwise stated, we used exponential learning rate decay, starting at $0.01$ and ending at $0.0001$.
We employ a hundredfold reduced learning rate for the first three epochs, as we find this helps set training of invertible networks on a stable track.

Where we use recursive coupling blocks, the number of neurons per hidden layer in the subnetworks is decreased by half with each level of recursion, down to a minimum of $1/8$ of the top level network size.
Each subnetwork consists of two hidden layers and ReLU activations.
Network parameters are initialized from a Gaussian distribution with $\sigma = 0.005$.

\subsubsection{UCI Data}

For the \textsc{Power} data set, models were scaled to a budget of $5 \cdot 10^5$ trainable parameters and we used a batch size of $1660$ (i.e.~1000 batches/epoch).

For the \textsc{Gas} data set, models were scaled to a budget of $5 \cdot 10^5$ trainable parameters and we used a batch size of $853$ (i.e.~1000 batches/epoch).

For the \textsc{Miniboone} data set, models were scaled to a budget of $2.5 \cdot 10^5$ trainable parameters and we used a batch size of $300$ (i.e.~100 batches/epoch).

\subsubsection{Lens Shapes}

We instantiate this data set with $10^{6}$ training and $10^{5}$ test samples and use a batch size of $10^{4}$.
All unconditional models were scaled to a budget of $1 \cdot 10^5$ trainable parameters, all conditional models to a budget of $4 \cdot 10^5$.

\subsubsection{Cross Shapes}

We instantiate this data set with $10^{6}$ training and $10^{5}$ test samples and use a batch size of $10^{4}$.
All unconditional models were scaled to a budget of $2 \cdot 10^6$ trainable parameters, all conditional models to a budget of $4 \cdot 10^6$.
For some of the deeper models shown in the trade-off section later on (16 and 32 blocks), we initialize the learning rate to $0.001$ to stabilize training.

\section{Shape Fitting Algorithm}
\label{sec:appendix-fitting}

In order to judge the quality of individual generated shapes, we compare to the best match among shapes from the ground truth distribution.
This means we have to find the best match in an automated fashion.
The way we do this differs slightly for the two types of shapes we used.

The algorithm for fitting a proper \emph{Lens} shape $\hat{S}$ -- parametrized by a tuple $\theta$ consisting of angle, scale and center -- to a generated one $S$ works as follows:
\begin{enumerate}
    \item find the two points $\mathbf{P}_0, \mathbf{P}_1 \in S$ which are furthest apart
    \item $\alpha =\ $ the angle of the line $\overrightarrow{\mathbf{P}_0 \mathbf{P}_1}$
    \item initialize shape parameters $\theta$ to $\bigl( \alpha, 2, \tfrac{1}{n} \sum_{j=1}^n \mathbf{S}_j \bigr)$
    \item use gradient descent to minimize $\mathcal{L} \bigl( \hat{S}(\theta), S \bigr)$, where
        \begin{linenomath*}
        \begin{align*}
            \mathcal{L} \bigl( \hat{S}(\theta), S \bigr) = \frac{1}{m} \sum_{i=1}^m \min_j D_{ij} + \frac{1}{n} \sum_{j=1}^n \min_i D_{ij} \\
            \text{with } \mathbf{D} \in \mathbb{R}^{m \times n} \text{ and } D_{ij} = \| \mathbf{\hat{S}}_i - \mathbf{S}_j \|^2_2
        \end{align*}
        \end{linenomath*}
    \item repeat for initialization with angle $\alpha' = \alpha + \pi$ (i.e.~the ``flipped'' lens) and keep the result with lower $\mathcal{L}$
\end{enumerate}
In other words, we minimize the average distance from each point in the fitted shape to the nearest point in the generated shape, and vice versa, starting from a strong heuristic guess for the correct angle.
Since the angle has such a strong effect on the fit and our initial guess should be reasonably good, we set a lower learning rate for the angle parameter during gradient descent.

The equivalent procedure for the more complex \emph{Cross} shapes data set requires a few changes to work well.
\begin{itemize}
    \item The shape $\hat{S}$ is parameterized by a tuple $\theta$ consisting of width, length and lateral shift of each of the two bars, as well as an angle $\alpha$ and the center of their crossing.
    \item The initial angle $\alpha$ is taken from the dominant straight line within $S$, as determined by RANSAC \cite{fischer1981ransac}.
    \item A general clean \emph{Cross} shape $\hat{S}$ is a twelve-sided polygon.
        We adapt the loss $\mathcal{L}$ from above by changing the first term to the average distance from each corner of the polygon $\hat{S}$ to the closest point in $S$,
        and the second term to the average distance from each point in $S$ to the closest line segment of $\hat{S}$.
    \item Because our \emph{Cross} shapes can have very diverse layouts, initializations with wrong lateral shift often get stuck in a bad local minimum.
    To counteract this, we repeat the optimization for each of the nine archetypes shown in \cref{fig:fit-plus-shapes} and keep the result where $\mathcal{L}$ is lowest.
\end{itemize}
With these changes we consistently get a good fit $\hat{S}$, provided the generated $S$ actually resembles a \emph{Cross} shape.

\section{Samples from Deeper Models}
\label{sec:appendix-deep-samples}

In \cref{fig:deep-samples-plus} we show qualitative results for deeper non-recursive models trained on the \emph{Cross} shapes data set, all under identical training conditions and at the same total parameter budget of $2M$.
Both visually and in terms of the $\x$ correlation matrix, recursive coupling with only 4 blocks is on par or better than non-recursive coupling with the same parameter budget spread across up to 32 blocks.

\section{Sensitivity of the Fourier Curve Parameterization}
\label{sec:appendix-sensitivity}

To demonstrate how quickly Fourier shapes deteriorate as soon as a single parameter goes ``out of sync'', we present \cref{fig:parameter-sensitivity}.
In each row, the same original shape is manipulated by adding an increasing offset ($0.1$ per column, \emph{left to right}) to one arbitrary Fourier parameter $x_i$.
It is clear that even minor mistakes in a single variable affect the entire shape.
The quality of generated samples thus relies heavily on modelling correlations between the variables correctly.

\section{Recursion Depth vs.~Number of Blocks}
\label{sec:appendix-depth-tradeoff}

We trained a number of models with varying recursion depth and number of blocks on the unconditional \emph{Cross} data set to investigate the trade-off between these hyperparameters.
The results are presented in \cref{tab:depth-comparison-fixed,tab:depth-comparison-128,tab:depth-comparison-512}.
A recursion depth of $0$ corresponds to standard \textsc{Real-NVP} coupling.
Note that for the latter two tables, we lift the total parameter budget and instead only set a fixed size for the coupling networks within each single block.

We find that in all cases, the best results both statistically and in terms of shape quality are achieved by networks with one or two levels of recursion.
Beyond that, the advantages of recursive coupling start to be outweighed by the parameter trade-off or the challenges of training more complex networks.
We also find that the deeper networks (i.e.~16 and 32 blocks) exhibit less stable training results, which we consider an argument in favor of using fewer, more expressive coupling blocks.

\section{Proof of Concept: \\ Bayesian filtering using HINT}
\label{sec:appendix-clv}

In the following, we compare HINT and a conditional adaptation of Real-NVP (\citet{ardizzone2018analyzing}, called INN for short) on a Bayesian filtering test case.
Filtering is a recursive Bayesian estimation task on hidden Markov models which requires to estimate the posterior of unobserved variables.
The posterior is then used as a prior for the next step, and subsequently updated by newly arriving observations.
Standard sampling algorithms for filtering require a Monte Carlo approximation of the analytic expression of the prior, which may dramatically deteriorate performance due to large variance.
HINT, on the other hand, does not need a functional form of the prior but only samples from it, which makes it an interesting candidate method for filtering.

As a task, we use the general predator-prey model, also termed \emph{competitive Lotka-Volterra equations},
which typically describes the interaction of $d$ species in a biological system over time:
\begin{linenomath*}
\begin{align}
    \frac{\partial}{\partial t} {x_i} &=  \beta_i x_i \left( 1 - \sum_{j=1}^{d} \alpha_{ij} x_j \right)
    \label{eq:lotka-volterra}
\end{align}
\end{linenomath*}
The undisturbed growth rate of species $i$ is given by $\beta_i$ (can be $<$ or $> 0$, growing or shrinking naturally),
and is further affected by the other species in a positive way ($\alpha_{ij} >0$, predator),
or in a negative way ($\alpha_{ij} <0$, prey).
The solutions to this system of equations can not be expressed analytically, which makes their prediction a challenging task. 
Additionally, we make the process stochastic, by adding random noise at each time step. 
Therefore, at each time step, noisy measurements of $x_1, x_2, x_3$ are observed, with the task to predict the remaining population $x_4$,
given the current and past observations.

We compare this to an implementation using a standard coupling block INN with the training procedure proposed by \citet{ardizzone2018analyzing}.
The results for a single example time-series are shown in \cref{fig:sequential}.
We find that the standard INN does not make meaningful predictions in the short- or mid-term, only correctly predicting the final dying off of population $x_4$.
HINT on the other hand is able to correctly model the entire sequence. Importantly, the modeled posterior distribution at each step correctly contracts over time, as more measurements are accumulated.
The experimental details are as follows:

\begin{centering}
    \resizebox{\linewidth}{!}{\includegraphics[]{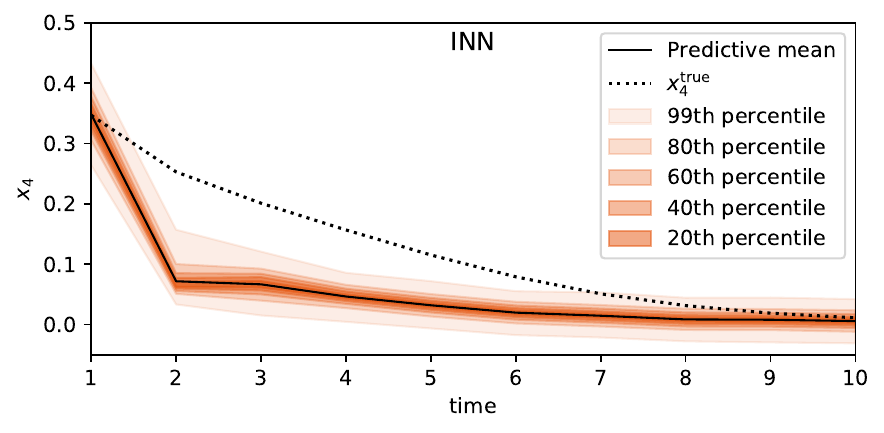} }
    \resizebox{\linewidth}{!}{\includegraphics[]{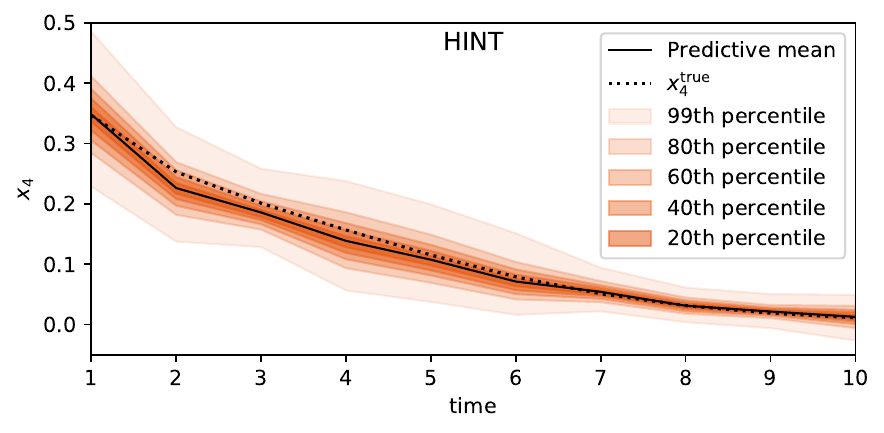} }
    \captionof{figure}{Contracting posterior for sequential Lotka-Volterra.}
    \label{fig:sequential}
    \vspace{2em}

    \resizebox{\linewidth}{!}{\includegraphics[]{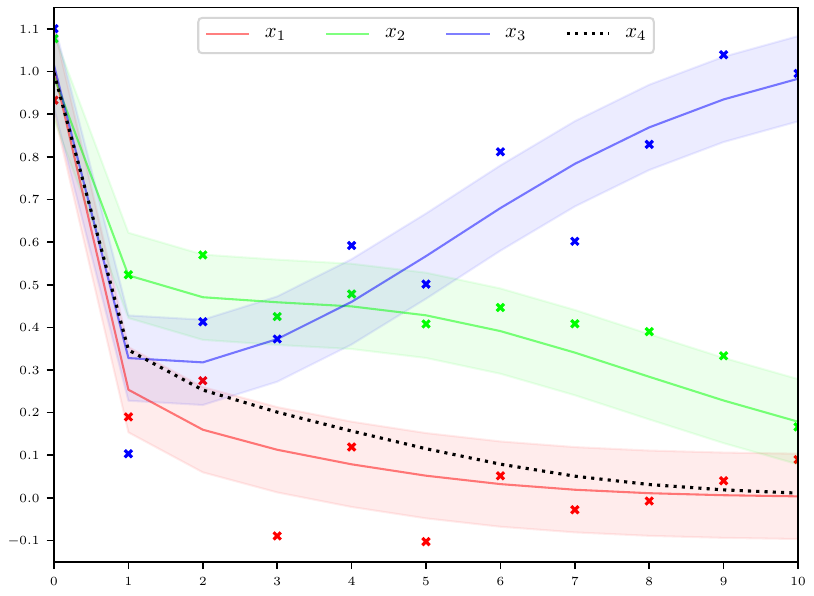} }%
    \captionof{figure}{Population development in competitive Lotka-Volterra model.
    $x_1$, $x_2$ and $x_3$ are only accessible via noisy measurements \emph{(small crosses)},
    the colored bands show the standard deviation of this noise.
    The task is to predict $x_4$ as the sequence develops.}
    \label{fig:populations}%
    \vspace{2em}
\end{centering}

\newpage

Parameters $\boldsymbol\alpha$ and $\boldsymbol\beta$ for the 4d competitive Lotka-Volterra model in \cref{eq:lotka-volterra}
were picked randomly as
\begin{linenomath*}
\begin{align*}
    \boldsymbol\beta &= \scalebox{0.9}{$\begin{bmatrix} 1.14143055 \\ 0.64270729 \\ 1.42981209 \\ 0.90620443 \end{bmatrix}$}
    \; \text{and} \\
    \boldsymbol\alpha &= \scalebox{0.9}{$\begin{bmatrix}
        0.78382338 & 1.26614888 & 1.25787652 & 0.80904295 \\
        1.00470891 & 0.32719451 & 1.34501072 & 1.29758381 \\
        1.28599724 & 0.39362355 & 0.89977679 & 1.00063551 \\
        1.12163602 & 1.08672758 & 1.39634746 & 0.53592833
    \end{bmatrix}$}, \\
    &\text{and the true initial population values $\x_0^\ast$ set to} \\
    \x_0^\ast &= \scalebox{0.9}{$\begin{bmatrix} 1.00471435 \\ 0.98809024 \\ 1.01432707 \\ 0.99687348 \end{bmatrix}$}.
\end{align*}
\end{linenomath*}

\Cref{fig:populations} shows the resulting development of the four populations over ten unit time steps, including noise on the observed values for $x_1$, $x_2$ and $x_3$.

Both the INN and the HINT network we trained consist of ten (non-recursive) coupling blocks with a total of $10^6$ trainable weights.
We used Adam to train both for $50$ epochs per time step with $64000$ training samples and a batch size of $500$.
The learning rate started at $10^{\sminus 2}$ and exponentially decayed to $10^{\sminus 3}$ (HINT) and $10^{\sminus 4}$ (INN), respectively.
Inference on $x_4$ begins with an initial guess at $t=0$ drawn from $\mathcal{N}(1,\tfrac{1}{10})$.

\begin{table*}[!b]
    \caption{Comparing the effect of more coupling blocks versus more levels of recursion within the blocks, while keeping the network parameter budget fixed at $2 \cdot 10^6$.}
    \label{tab:depth-comparison-fixed}
    \begin{center}\begin{small}\begin{sc}
    \begin{tabular}{l@{\quad}l@{\qquad\quad}
        S[detect-weight,mode=text,table-format=1.4]@{\qquad\quad}
        S[detect-weight,mode=text,table-format=1.4]@{\qquad\quad}
        S[detect-weight,mode=text,table-format=1.4]@{\qquad\quad}
        S[detect-weight,mode=text,table-format=1.4]}
    \toprule
    {Recursion depth} & & {0} & {1} & {2} & {3} \\
    \midrule
    \multirow{4}{*}[0pt]{4 blocks}  & LL $\uparrow$       & 3.419 & \B 3.680 & 3.663 & 3.666 \\
                                    & IoU $\uparrow$      & 0.314 & \B 0.857 & 0.823 & 0.835 \\
                                    & H-dist $\downarrow$ & 0.236 & \B 0.066 & 0.077 & 0.071 \\
                                    & Corr $\downarrow$   & 0.0454 & \B 0.0022 & 0.0027 & 0.0026 \\
    \midrule
    \multirow{4}{*}[0pt]{8 blocks}  & LL $\uparrow$       & 3.329 & 3.592 & 3.593 & \\
                                    & IoU $\uparrow$      & 0.588 & 0.802 & 0.817 & \\
                                    & H-dist $\downarrow$ & 0.138 & 0.085 & 0.083 & \\
                                    & Corr $\downarrow$   & 0.0064 & 0.0030 & 0.0028 & \\
    \cmidrule(r{2em}){1-5}
    \multirow{4}{*}[0pt]{16 blocks} & LL $\uparrow$       & 3.530 & 3.623 & & \\
                                    & IoU $\uparrow$      & 0.778 & 0.841 & & \\
                                    & H-dist $\downarrow$ & 0.101 & 0.073 & & \\
                                    & Corr $\downarrow$   & 0.0035 & 0.0026 & & \\
    \cmidrule(r{2em}){1-4}
    \multirow{4}{*}[0pt]{32 blocks} & LL $\uparrow$       & 3.578 & & & \\
                                    & IoU $\uparrow$      & 0.836 & & & \\
                                    & H-dist $\downarrow$ & 0.082 & & & \\
                                    & Corr $\downarrow$   & 0.0029 & & & \\
    \cmidrule[1pt](r{2em}){1-3}
    \end{tabular}
    \end{sc}\end{small}\end{center}
    \vskip -0.1in
\end{table*}

\begin{table*}[tb]
    \caption{Comparing the effect of more coupling blocks versus more levels of recursion within the blocks, while affording internal networks a width of $128$ neurons, halved at every recursion level.}
    \label{tab:depth-comparison-128}
    \begin{center}\begin{small}\begin{sc}
    \begin{tabular}{l@{\quad}l@{\qquad\quad}
        S[detect-weight,mode=text,table-format=1.4]@{\qquad\quad}
        S[detect-weight,mode=text,table-format=1.4]@{\qquad\quad}
        S[detect-weight,mode=text,table-format=1.4]@{\qquad\quad}
        S[detect-weight,mode=text,table-format=1.4]}
    \toprule
    {Recursion depth} & & {0} & {1} & {2} & {3} \\
    \midrule
    \multirow{4}{*}[0pt]{4 blocks}  & LL $\uparrow$       & 3.496 & 3.595 & 3.612 & 3.614 \\
                                    & IoU $\uparrow$      & 0.713 & 0.806 & 0.813 & 0.804 \\
                                    & H-dist $\downarrow$ & 0.115 & 0.090 & 0.085 & 0.082 \\
                                    & Corr $\downarrow$   & 0.0032 & 0.0025 & 0.0024 & 0.0023 \\
    \midrule
    \multirow{4}{*}[0pt]{8 blocks}  & LL $\uparrow$       & 3.623 & \B 3.664 & 3.651 & \\
                                    & IoU $\uparrow$      & 0.822 & \B 0.864 & 0.860 & \\
                                    & H-dist $\downarrow$ & 0.081 & \B 0.065 & 0.067 & \\
                                    & Corr $\downarrow$   & 0.0023 & \B 0.0018 & \B 0.0018 & \\
    \cmidrule(r{2em}){1-5}
    \multirow{4}{*}[0pt]{16 blocks} & LL $\uparrow$       & 3.631 & 3.613 & & \\
                                    & IoU $\uparrow$      & 0.839 & 0.819 & & \\
                                    & H-dist $\downarrow$ & 0.074 & 0.080 & & \\
                                    & Corr $\downarrow$   & 0.0020 & 0.0022 & & \\
    \cmidrule(r{2em}){1-4}
    \multirow{4}{*}[0pt]{32 blocks} & LL $\uparrow$       & 3.612 & & & \\
                                    & IoU $\uparrow$      & 0.814 & & & \\
                                    & H-dist $\downarrow$ & 0.086 & & & \\
                                    & Corr $\downarrow$   & 0.7490 & & & \\
    \cmidrule[1pt](r{2em}){1-3}
    \end{tabular}
    \end{sc}\end{small}\end{center}
    \vskip -0.1in
\end{table*}

\begin{table*}[tb]
    \caption{Comparing the effect of more coupling blocks versus more levels of recursion within the blocks, while affording internal networks a width of $512$ neurons, halved at every recursion level.}
    \label{tab:depth-comparison-512}
    \begin{center}\begin{small}\begin{sc}
    \begin{tabular}{l@{\quad}l@{\qquad\quad}
        S[detect-weight,mode=text,table-format=1.4]@{\qquad\quad}
        S[detect-weight,mode=text,table-format=1.4]@{\qquad\quad}
        S[detect-weight,mode=text,table-format=1.4]@{\qquad\quad}
        S[detect-weight,mode=text,table-format=1.4]}
    \toprule
    {Recursion depth} & & {0} & {1} & {2} & {3} \\
    \midrule
    \multirow{4}{*}[0pt]{4 blocks}  & LL $\uparrow$       & 3.620 & 3.685 & 3.685 & 3.681 \\
                                    & IoU $\uparrow$      & 0.846 & 0.837 & 0.825 & 0.810 \\
                                    & H-dist $\downarrow$ & 0.082 & 0.069 & 0.068 & 0.071 \\
                                    & Corr $\downarrow$   & 0.0023 & 0.0025 & 0.0023 & 0.0025 \\
    \midrule
    \multirow{4}{*}[0pt]{8 blocks}  & LL $\uparrow$       & 3.669 & 3.716 & \B 3.722 & \\
                                    & IoU $\uparrow$      & 0.863 & 0.883 & \B 0.888 & \\
                                    & H-dist $\downarrow$ & 0.064 & 0.053 & \B 0.052 & \\
                                    & Corr $\downarrow$   & 0.0020 & 0.0018 & \B0.0017 & \\
    \cmidrule(r{2em}){1-5}
    \multirow{4}{*}[0pt]{16 blocks} & LL $\uparrow$       & 3.702 & 3.685 & & \\
                                    & IoU $\uparrow$      & 0.882 & 0.878 & & \\
                                    & H-dist $\downarrow$ & 0.055 & 0.060 & & \\
                                    & Corr $\downarrow$   & \B 0.0017 & \B 0.0017 & & \\
    \cmidrule(r{2em}){1-4}
    \multirow{4}{*}[0pt]{32 blocks} & LL $\uparrow$       & 3.654 & & & \\
                                    & IoU $\uparrow$      & 0.813 & & & \\
                                    & H-dist $\downarrow$ & 0.072 & & & \\
                                    & Corr $\downarrow$   & 0.0021 & & & \\
    \cmidrule[1pt](r{2em}){1-3}
    \end{tabular}
    \end{sc}\end{small}\end{center}
    \vskip -0.1in
\end{table*}

\clearpage
\section{Acknowledgments}
Jakob Kruse was supported by Informatics for Life funded by the Klaus Tschira Foundation.
Gianluca Detommaso was supported by the EPSRC Centre for Doctoral Training in Statistical Applied Mathematics at Bath (EP/L015684/1).

\bibliography{references}

\end{document}